\documentclass{article}

\usepackage[final]{neurips_2022}
\usepackage[utf8]{inputenc} 
\usepackage[T1]{fontenc}    
\usepackage{hyperref}       
\usepackage{url}            
\usepackage{booktabs}       
\usepackage{amsfonts}       
\usepackage{nicefrac}       
\usepackage{microtype}      
\usepackage{xcolor}  

\def\BibTeX{{\rm B\kern-.05em{\sc i\kern-.025em b}\kern-.08emT\kern-.1667em\lower.7ex\hbox{E}\kern-.125emX}}

\usepackage{color}
\usepackage{graphicx}
\usepackage{multicol}
\usepackage{multirow}
\usepackage{array}
\usepackage{xspace}
\usepackage{enumitem}
\setlist[itemize]{leftmargin=*}
\setlist[enumerate]{leftmargin=*}
\usepackage{amsmath}
\usepackage[font=small,labelfont=bf]{caption}
\usepackage{tablefootnote}

\newcommand{\notationFont}{\mathsf}

\newcommand{\action}{\notationFont{a}}

\newcommand{\staterep}{\notationFont{s}}

\newcommand{\reward}{\notationFont{r}}

\newcommand{\datasetFont}{\textsf}
\newcommand{\gridone}{\datasetFont{CityFlow1x1}\xspace}
\newcommand{\realone}{\datasetFont{Cologne1x1}\xspace}
\newcommand{\realthree}{\datasetFont{Cologne1x3}\xspace}
\newcommand{\gridfour}{\datasetFont{Grid4x4}\xspace}
\newcommand{\hzfour}{\datasetFont{HangZhou4x4}\xspace}

\newcommand{\ours}{\textit{LibSignal}\xspace}
\newcommand{\ie}{\textit{i.e.}\xspace}

\newcommand{\methodFont}{\textit}
\newcommand{\ft}{\methodFont{FixedTime}\xspace}
\newcommand{\sotl}{\methodFont{SOTL}\xspace}
\newcommand{\maxp}{\methodFont{MaxPressure}\xspace}
\newcommand{\dqn}{\methodFont{IDQN}\xspace}
\newcommand{\frap}{\methodFont{FRAP}\xspace}
\newcommand{\maddpg}{\methodFont{MAPG}\xspace}
\newcommand{\ppo}{\methodFont{IPPO}\xspace}
\newcommand{\presslight}{\methodFont{PressLight}\xspace}
\newcommand{\mplight}{\methodFont{MPLight}\xspace}
\newcommand{\colight}{\methodFont{CoLight}\xspace}

\newcommand{\rebuttal}{}
\renewcommand{\thefootnote}{\fnsymbol{footnote}}

\title{LibSignal: An Open Library for Traffic Signal Control}

\author{%
  Hao Mei\footnotemark[1] \ $^{\dag}$, Xiaoliang Lei\footnotemark[1] \ $^{\ddag}$, Longchao Da$^{\dag}$, Bin Shi$^ {\ddag}$, Hua Wei\footnotemark[4] \ $^{\dag}$ \\
    $^{\dag}$New Jersey Institute of Technology, $^{\ddag}$Xi'an Jiaotong University\\
  hm467@njit.edu, shawlenleo@stu.xjtu.edu.cn, ld49@njit.edu, \\
  shibin@xjtu.edu.cn, hua.wei@njit.edu \\
}

\begin{document}

\maketitle
\renewcommand{\thefootnote}{\fnsymbol{footnote}}
\footnotetext[1]{Authors contributed equally.}
\footnotetext[4]{Corresponding author.}
\renewcommand{\thefootnote}{\arabic{footnote}}

\begin{abstract}
This paper introduces a library for cross-simulator comparison of reinforcement learning models in traffic signal control tasks. This library is developed to implement recent state-of-the-art reinforcement learning models with extensible interfaces and unified cross-simulator evaluation metrics. It supports commonly-used simulators in traffic signal control tasks, including Simulation of Urban MObility(SUMO) and CityFlow, and multiple benchmark datasets for fair comparisons. We conducted experiments to validate our implementation of the models and to calibrate the simulators so that the experiments from one simulator could be referential to the other. Based on the validated models and calibrated environments, this paper compares and reports the performance of current state-of-the-art RL algorithms across different datasets and simulators. This is the first time that these methods have been compared fairly under the same datasets with different simulators. 
\end{abstract}

\section{Introduction}
\label{sec:intro}
Traffic signals coordinate the traffic movements at the intersection, and a smart traffic signal control algorithm is the key to transportation efficiency. Traffic signal control remains an active research topic because of the high complexity of the problem. The traffic situations are highly dynamic and thus require traffic signal plans to adjust to different situations. People have recently started investigating reinforcement learning (RL) techniques for traffic signal control. Several studies have shown the superior performance of RL techniques over traditional transportation approaches~\cite{wei2018intellilight,wei2019presslight,xu2021hierarchically,oroojlooy2020attendlight,ma2020feudal}. The biggest advantage of RL is that it directly learns how to take the next actions by observing the feedback from the environment after previous actions.

In literature, a number of traffic signal control methods have been proposed~\cite{yau2017survey,wei2019survey}, and it has attracted much attention to facilitate the implementation or use of these proposed methods. However, as shown in Table~\ref{tab:papers}, current methods are distributed among different simulators and datasets. As we will show later in this paper, datasets, simulators, and even evaluation metrics var\rebuttal{y} the performance for the same algorithm. In addition, reinforcement learning is also sensitive to hyper-parameters. All these make it difficult for new traffic signal control methods to ensure effective and uniform improvement. Therefore, there is an urgent need for a cross-platform, unified process with extensible code base that supports multiple models.


This paper presents a unified, flexible, and comprehensive traffic signal control library named \ours. Our library is implemented based on PyTorch and includes all the necessary steps or components related to traffic signal control into a systematic pipeline. We consider two mainstream simulators, SUMO and CityFlow, and provide various datasets, models, and utilities to support data preparation, environment calibration, model instantiation, and performance evaluation for the two kinds of simulators.

\textbf{Contr\rebuttal{i}butions}: To the best of our knowledge, \ours is the first open-source library that provide\rebuttal{s} benchmarking results for traffic signal control methods across various datasets and simulators.
 The main features of \ours can be summarized in three aspects:
\begin{itemize}
    \vspace{-.5em}
    \item Unified: \ours builds a systematic pipeline to implement, use and evaluate traffic signal control models in a unified platform. We design cross-simu\rebuttal{la}tor data configuration, unified model instantiation interfaces, and standardized evaluation procedures.
    \vspace{-.2em}
    \item Comprehensive: \rebuttal{10} models covering two traffic simulators have been reproduced to form a comprehensive model warehouse. Meanwhile, \ours collects 9 commonly used datasets \rebuttal{from} different sources, makes them compatible for both simulators, and implements a series of commonly used evaluation metrics and strategies for performance evaluation. 
    \vspace{-.2em}
    \item Extensible: \ours enables a modular design of different components, allowing users to flexibly insert customized components into the library. It also has an OpenAI Gym interface~\cite{gym2016} which allows easy deployment of standard RL algorithms. 
\end{itemize}
\vspace{-.5em}
\textbf{What \ours isn’t}: Despite the ability to train and test across differen\rebuttal{t} simulators, \ours does not claim the performances of the same model on different simulators are identical. There
are some differences of internal mechanisms between different simulators, for example, vehicle's maneuver behaviors, and our emphasis is on the relative performance of compatible policies we provide. This makes \ours a possible testbed for Sim-to-Real transfer~\cite{zhao2020sim, peng2018sim}, which is not covered by this paper.

\begin{table}[t!]
\tiny
\centering
\caption{Summary of the state-of-the-art models (a partial list), sorted by citations from Google Scholar by 2022/06/08. Full list can be found in~\url{https://darl-libsignal.github.io/}.}
\label{tab:papers}
\begin{tabular}{|c|c|c|c|c|c|}
\hline
Method       & Venue        & Cite & Simulator & Dataset (* means open accessed)  & Main Metrics                                     \\ \hline

\rebuttal{IntelliLight}~\cite{wei2018intellilight} & KDD 2018     & 321      & SUMO      & Jinan 1x1                                                                                     & \begin{tabular}[c]{@{}c@{}}Travel Time, speed, queue length, \\ approximated delay\end{tabular}  \\ 
\hline

\rebuttal{IDQN} \rebuttal{~\cite{{https://doi.org/10.48550/arxiv.1905.04716}}} & \rebuttal{arXiv 2019}     & \rebuttal{53}      & SUMO      & \rebuttal{LA 1x4*, Jinan 1x1, Hangzhou 1x1}                                                                                    & \begin{tabular}[c]{@{}c@{}}\rebuttal{Travel Time}\end{tabular}  \\ \hline
MAPG~\cite{chu2019multi}     & TITS 2019    & 308      & SUMO      & Grid 4x4, Monaco*                                                                                  & Approximated delay, queue length   \\ \hline
CoLight~\cite{wei2019colight}  & CIKM 2019    & 116      & CityFlow  & \begin{tabular}[c]{@{}c@{}}Hangzhou 4x4*, Jinan 3x4*, \\ Manhattan 3x27*\end{tabular}            & Travel Time   \\ \hline
MPLight~\cite{chen2020toward}    & AAAI 2020    & 98       & CityFlow  & Grid 4x4, Manhattan                                                                           & Travel time, throughput    \\ \hline
PressLight~\cite{wei2019presslight}   & KDD 2019     & 93       & CityFlow  & Jinan 1x3, State College*, Mahattan*                                                             & Travel time     \\ \hline
FRAP~\cite{zheng2019learning}      & CIKM 2019    & 67       & CityFlow  & Hangzhou 1x1*, Atlanta 1x5*                                                                     & Travel Time    \\ \hline
MetaLight~\cite{zang2020metalight}   & AAAI 2020    & 43       & CityFlow  & \begin{tabular}[c]{@{}c@{}}Hangzhou 1x1*, Atlanta 1x5*, \\ Hangzhou 4x4*, Jinan 3x4*\end{tabular} & Travel time  \\ \hline
DemoLight~\cite{xiong2019learning}  & CIKM 2019    & 22       & CityFlow  & Hangzhou 1x1*                                                                                  & Travel Time  \\ \hline
FMA2C ~\cite{ma2020feudal}    & AAMAS 2020   & 16       & SUMO      & 4x4 Grid, Monaco*                                                                              & Queue length, throughput, delay    \\ \hline
TPG ~\cite{rizzo2019time}     & KDD 2019     & 15       & SUMO      & Roundabout                                                                                    & \begin{tabular}[c]{@{}c@{}}Queue length, waiting time, \\ throughput, speed\end{tabular}   \\ \hline
AttendLight~\cite{oroojlooy2020attendlight}  & NeurIPS 2020 & 12       & CityFlow  & Hangzhou 4x4*, Atlanta 1x5*                                                                     & Travel time  \\ \hline
HiLight~\cite{xu2021hierarchically}   & AAAI 2021    & 12       & CityFlow  & \begin{tabular}[c]{@{}c@{}}Hangzhou 4x4*, Jinan 3x4*, \\ Manhattan 3x27*, Shenzhen*\end{tabular}  & Travel time, throughput  \\ \hline
IG-RL~\cite{devailly2021ig}   & TITS 2021    & 11       & SUMO      & Manhattan                                                                                     & Approximated delay  \\ \hline
ExplainPG~\cite{rizzo2019reinforcement}  & ITSC 2019    & 11       & SUMO      & Roundabout                                                                                    & Waiting time, throughput   \\ \hline
RACS-R~\cite{wang2021traffic}     & TITS 2021    & 6        & SUMO      & Monaco*                                                                                        & Waiting time, queue length     \\ \hline
GeneraLight~\cite{zhang2020generalight}  & CIKM 2020    & 4        & CityFlow  & \begin{tabular}[c]{@{}c@{}}Hangzhou 1x1*, Atlanta 1x5*, \\ Hangzhou 4x4*\end{tabular}            & Travel time   \\ \hline
OP-TSC~\cite{yen2020deep}       & ITSC 2020    & 4        & SUMO      & synthetic                                                                                     & Delay   \\ \hline
DFC~\cite{raeis2021deep}    & ITSC 2021    & 2        & SUMO      & synthetic                                                                                     & Waiting time   \\ \hline
DynSTGAT~\cite{wu2021dynstgat}  & CIKM 2021    & 0        & CityFlow  & Hangzhou 4x4*, Jinan 3x4*, Grid 4x4*                                                             & Travel time, throughput  \\ \hline
EMV~\cite{cao2022gain}      & TITS 2022    & 0        & SUMO      & Hangzhou 4x4*                                                                                  & Waiting time, queue length \\ \hline
\end{tabular}
\end{table}

\section{Background}


\subsection{Reinforcement Learning for Traffic Signal Control}
\paragraph{Problem formulation}
We now introduce the general setting of the RL-based traffic signal control problem, in which the traffic signals are controlled by an RL agent or several RL agents. The environment is the traffic conditions on the roads, and the agents control the traffic signals\rebuttal{' phases}. At each time step $t$, a description of the environment (e.g., signal phase, waiting time of cars, queue length of cars, and positions of cars) will be generated as the state $\staterep^t$. The agents will predict the next actions $\action^{t}$ to take that maximize the expected return,
where the action\rebuttal{ of a} single intersection could be changing to a certain phase. The actions $\action^t$ will be executed in the environment, and a reward $\reward^{t}$ will be generated, where the reward could be defined on the traffic conditions of the intersections. 



\vspace{-1em}
\paragraph{Basic components of RL-based traffic signal control}
A key question for RL is how to formulate the RL setting, i.e., the reward, state and action definition. For more discussions on the reward, state, and actions, we refer interested readers to~\cite{yau2017survey,rasheed2020deep, wei2021recent}. There are \rebuttal{three} main components to formulate the problem under the framework of RL:

\begin{itemize}
\vspace{-1em}
\item Reward design. As RL is learning to maximize a numerical reward, the choice of reward determines the direction of learning. A typical reward definition for traffic signal control is one factor or a weighted linear combination of several components such as queue length, waiting time and delay.
\vspace{-.5em}
\item State design. State captures the situation on the road and converts it to values. Thus the choice of states should sufficiently describe the environment.  The state features queue length, the number of cars, waiting time, and the current traffic signal phase. Images of vehicles' positions on the roads can also be considered into the state.
\vspace{-.5em}
\item Selection of action scheme. Different action schemes also have influences on the performance of traffic signal control strategies. For example, if the action of an agent is \emph{acyclic}, \ie, ``which phase to change to'', the traffic signal will be \rebuttal{more} flexible than \rebuttal{a} \emph{cyclic} action, \ie, ``keep current phase or change to the next phase in a cycle''.  
\end{itemize}

\subsection{Difficulties in Evaluation}
In practice, the evaluation of traffic signal control methods could be largely influenced by simulation settings, including the evaluation metrics and simulation environments.
\vspace{-1em}
\paragraph{Evaluation metrics}
Various measures have been proposed to quantify the efficiency of the intersection from different perspectives, including the average delay of vehicles, the average queue length in the road network, the average travel time of all vehicles, and the throughput of the road network. Signal induced \emph{delay} is another widely used metric, and previous work suggested real-time approximation as the difference between the vehicle’s current speed and the maximum speed limit over all vehicles. But as we will show in Section~\ref{sec:exp:overall}, this approximated delay is not reflecting the actual delay. \emph{Queue length} is another mostly used metrics~\cite{wei2018intellilight}, while different definitions of a "queuing" state of a vehicle could largely influence the performance of the same method. In comparison, \emph{travel time} and \emph{throughput} are more robust to ad-hoc definitions and approximations. As we will show later, with the same experimental setting, the performance of the same method could be different under different \rebuttal{metric}, and we aim to provide as comprehensive and flexible metrics as possible in this paper to benchmark methods with a comprehensive view.
\vspace{-1em}
\paragraph{Simulation environments}
Since deploying and testing traffic signal control strategies in the real world involve high cost and intensive labor, simulation is a useful alternative before actual implementation. Different choice\rebuttal{s} of simulator could lead to different evaluation performances.

Currently, there are two representative open-source microscopic simulators: \emph{Simulation of Urban MObility (SUMO)}\footnote{http://sumo.sourceforge.net}~\cite{sumo2018} and \emph{CityFlow}\footnote{https://cityflow-project.github.io/}~\cite{zhang2019cityflow}. SUMO is widely accepted in the transportation community and is a reasonable testbed choice. 
Compared with SUMO, CityFlow is a simulator optimized for reinforcement learning with faster simulation, while it is not widely used by the transportation field yet. 

Because of these different simulation environments, methods adopted by different simulators in their original paper\rebuttal{s} are hard to evaluate. As we will show later, methods perform differently under different well-calibrated simulators, and the efficiency of the training process is also different under different simulators. This paper, for the first time, compares the performances of the same model under the same traffic datasets under different simulators.

\subsection{Existing Libraries and Tools}
~\cite{websiteTSC} is an open-source library that provides a bunch of RL-based traffic signal control methods with traffic datasets \emph{only} on CityFlow~\cite{zhang2019cityflow}. Flow~\rebuttal{\cite{kheterpal2018flow}} and RESCO~\cite{ault2021reinforcement} are reinforcement learning frameworks that can support the design and experimentation of traffic signal control methods \emph{only} on SUMO~\cite{sumo2018}. TSLib~\rebuttal{\cite{tslib2021}} is another library that could work under both SUMO and CityFlow, yet it has limited extensibility: (1) it is difficult to deploy standard RL algorithms since it does not have an OpenAI Gym interface; (2) there \rebuttal{are} no benchmarking datasets that \rebuttal{work} across both simulators, which makes it challenging to help determine which algorithm results in state-of-the-art performance.
\section{\ours Toolkit}
\label{sec:lib}
We propose \ours library integrating different influential traffic flow simulators and denote it as a standard RL traffic control testbed. The primary purposes of this standard testbed are:
\begin{itemize}[leftmargin=3em]
    \item[1.]
    Provide a converter to transform configurations including road net\rebuttal{works} and traffic flow files across different simulators, enabling comparisons between different algorithms originally conducted in different simulators.
    \item[2.]
    A standardized implementation of state-of-the-art RL-based and traditional traffic control algorithms.
    \item[3.] 
    A cross-simulator environment provides highly unified functions to interact with different baseline\rebuttal{s} or user-defined models and supports performance comparisons among them. 
\end{itemize}

\begin{figure*}[h!]
\centering
\includegraphics[width=.9\linewidth]{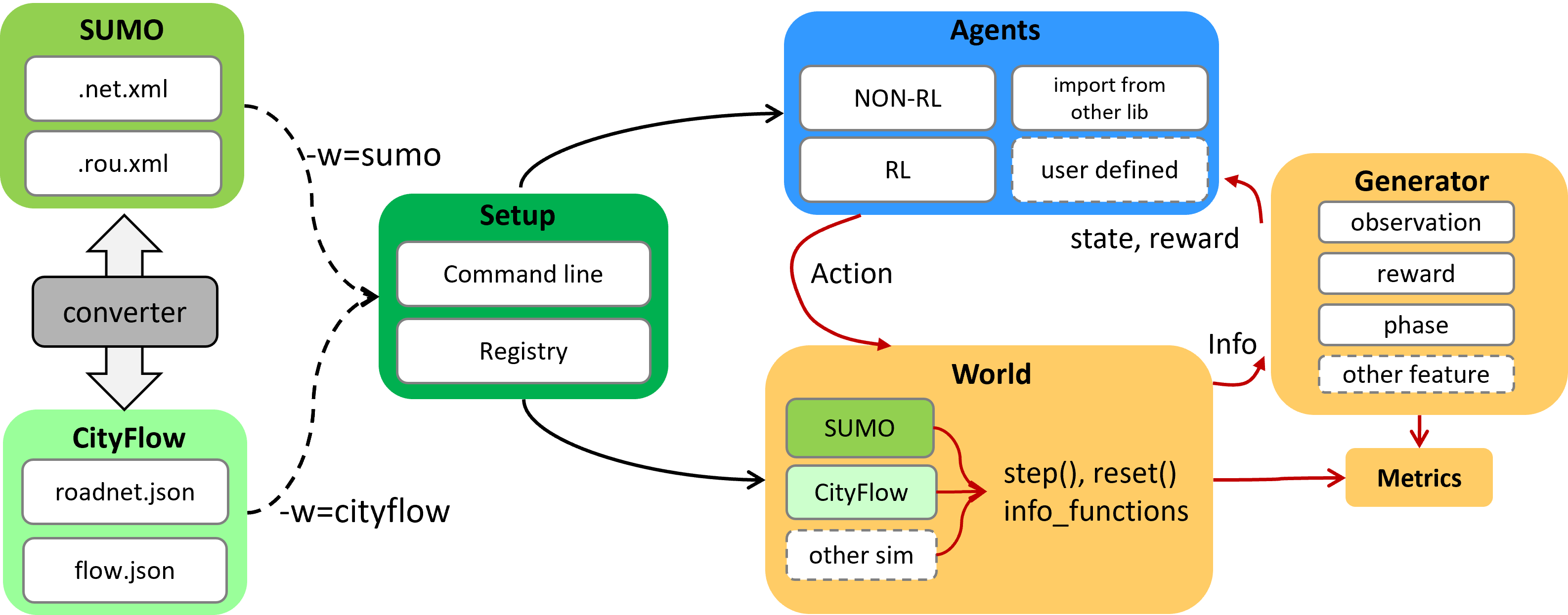}
\caption{Overall framework of \ours}
\label{fig:Overall framework}
\end{figure*}

\ours is open source and free to use/modify under the GNU General Public License 3. The code is built on top of GeneraLight~\cite{zhang2020generalight} and is available on Github at ~\url{https://darl-libsignal.github.io/}.
\rebuttal{The embedded traffic datasets are distributed with their own licenses from~\cite{websiteTSC} and~\cite{ault2021reinforcement}, whose licenses are under the GNU General Public License 3. SUMO is licensed under the EPL 2.0, and CityFlow is under Apache 2.0.} The overall framework of \ours is presented in Figure~\ref{fig:Overall framework}, and the implementation details will be introduced in the following sections.

\subsection{Data Preparation}
To enable fair comparison, \ours preprocesses comprehensive datasets \rebuttal{making it} runnable under different simulators. Users can easily choose to specify datasets and simulators for their experiments.  
\vspace{-2em}
\paragraph{Comprehensive datasets}
By surveying the recent literature\rebuttal{s} on traffic prediction, we select\rebuttal{ed} 225 representative or survey papers (more details can be found in Table~\ref{tab:papers}). We collected all the open datasets used by these papers and kept 9 datasets according to the factors of popularity, which can cover 65\% papers of our reproduced model list and all the two simulators \ours supports. To directly use these datasets in \ours, we have converted all the 9 datasets into the format of atomic files, and \rebuttal{provided} the conversion tools for new datasets. Please refer to our GitHub page for dataset statistics, preprocessed copies, and conversion tools at \rebuttal{~\url{https://darl-libsignal.github.io/.}}

\vspace{-1em}
\paragraph{Cross-simulator atomic files}
To make \rebuttal{the }experimental configuration adaptive across different simulators, we consider two basic units called ``atomic files" that can map to the different simulation environments. 1) \textit{Road network file} stores the basic structure of a traffic network consisting of road, lane, and traffic light information. The atomic file under the SUMO environment is in the format of \texttt{.net.xml} while in CityFlow it's \texttt{.json}. 2) \textit{Traffic flow file} stores the vehicles information and is in \texttt{.rou.xml} and \texttt{.json} format in SUMO and CityFlow respectively.
To make experiments comparable among different simulators, we also provide a \texttt{converter.py} tool to convert basic atomic files between different simulators. For example, it takes in \textit{Road network file} and \textit{Traffic flow file} from the source simulator and generates new files in the target simulator's formation, which could later be used in experiments. Figure~\ref{fig:tf-th} shows the converted network between different simulators. 

\subsection{Traffic Signal Control Environment}
Once the necessary parameters have been set up for simulation and agents, we can start a traffic light control task experiment. 
The \texttt{World} environment is highly homogeneous across different simulators and could provide unified interfaces to communicate with different agents.
\vspace{-.5em}
\paragraph{Homogeneous world}
In \ours, \texttt{World} module provides the basic information from different simulators in unified interfaces compatible with OpenAI Gym~\cite{gym2016}, which could later be utilized for interacting between different simulators and \texttt{Agent}. In the \texttt{World} class, we provide an \texttt{info\_functions} object inside to help retrieve information from different simulator environments and update information after each simulator \rebuttal{performs a step}. \rebuttal{The \texttt{info\_functions} contain \texttt{state} information including \texttt{lane\_count}, \texttt{lane\_waiting\_count}, \texttt{lane\_waiting\_time count}, \texttt{pressure}, \texttt{phase} ,and \texttt{metrics} including \texttt{throughput}, \texttt{average\_travel\_time}, \texttt{lane\_delay}, \texttt{lane\_vehicles.}} These \texttt{info\_functions} will later be called by \texttt{Generator} class and pass information into \texttt{Agent}. \texttt{step()} function is another common function shared between different \texttt{World} classes. It takes in actions returned from \texttt{Agent} class and \rebuttal{passes} them into the simulator for next step execution. \rebuttal{And \texttt{action} is either sampled from action space for exploration or calculated from the model after optimization. Generally, the action space contains eight phases. However, in highly heterogenous traffic structures, the action space may differ and is provided by the simulators whose action parameters are taken from configuration files.} 
\vspace{-.5em}
\paragraph{Unified interfaces}
\ours provide\rebuttal{s} unified interfaces to process common information with \texttt{Generator} module and \texttt{Metrics} module.
For lane level information, including state, reward, phase, and other lane level metrics, we provide \texttt{Generator} module, which could interact with different \texttt{World} classes and then sort and pass information to different \texttt{Agent} classes. The state, reward, and phase information will later be utilized by \texttt{Agent} module to train their models or decide next step actions and feedback to \texttt{World} module. At the same time, the other lane level metrics, including queue length and lane delay, will be passed to \texttt{Metrics} module for model evaluation. \ours currently support\rebuttal{s} four metrics: the average delay of vehicles (\emph{delay}), the average queue length in the road network (\emph{queue}), the average travel time of all vehicles (\emph{travel time}), and the throughput of the road network (\emph{throughput}). Their detailed calculation can be found in Appendix.



\subsection{Comprehensive Models}
\ours implements three baseline controllers and seven RL-based controllers covering Q-learning and Actor-Critic methods, as is shown in Table~\ref{tab:agents}. These methods can also be integrated with \rebuttal{existing} RL implementation packages and customized on their state, action, and reward design.

    
    
    
    
    
    

\begin{table}[tbh]
\tiny
\centering
\caption{Detailed design of implemented models in \ours.}
\label{tab:agents}
\resizebox{\columnwidth}{!}{
\begin{tabular}{|c|c|c|c|c|l|}
\hline
Agent       & State & Action  & Reward & Method          & Description                                                                                                                                                              \\ \hline
FixedTime     & -     & Cyclic  & -      & Non-RL          & \begin{tabular}[c]{@{}l@{}}This agent gives a predefined time duration \\ and phase order.\end{tabular}                                                                  \\ \hline
SOTL        & -     & Acyclic & -      & Non-RL          & \begin{tabular}[c]{@{}l@{}}This agent selects the phase among all to \\ maximize the pressure  calculated from the \\ upstream and downstream queue length.\end{tabular} \\ \hline
MaxPressure & -     & Acylic  & -      & Non-RL          & \begin{tabular}[c]{@{}l@{}}This agent determines next phase by \\ considering competitive phases.\end{tabular}                                                           \\ \hline
IDQN       &    lane vehicle count, phase     & Acylic  &  lane waiting vehicle  count   & Q-Learning  & \begin{tabular}[c]{@{}l@{}}This agent determines each intersection's \\ action with its own intersection information.\end{tabular}                                       \\ \hline
CoLight     &   lane vehicle count, phase   & Acylic  &    lane waiting vehicle count    & Q-Learning  & \begin{tabular}[c]{@{}l@{}}This agent considers neighbor intersections' \\ cooperation through  graph attention networks.\end{tabular}                                   \\ \hline
PressLight  &   lane vehicle count, phase    & Acylic  &   pressure     & Q-Learning  & \begin{tabular}[c]{@{}l@{}}This agent coordinates traffic signals by \\ learning MaxPressure.\end{tabular}                                                               \\ \hline
IPPO        &   lane vehicle count, phase    & Acylic  &    lane vehicle waiting time count    & Actor-Critic & \begin{tabular}[c]{@{}l@{}}This agent is imported from \texttt{pfrl} with \\proximal policy optimization.\end{tabular}                                                     \\ \hline
MAPG        &    lane vehicle count   & Acylic  &   lane waiting vehicle count     & Actor-Critic & \begin{tabular}[c]{@{}l@{}}This agent optimizes agent control policy \\ with muti-agent policy gradient method.\end{tabular}                                        \\ \hline
\rebuttal{FRAP}         &  \rebuttal{lane vehicle count, phase}    &  \rebuttal{Acylic} &    \rebuttal{lane waiting vehicle count}    & \rebuttal{Q-Learning} & \begin{tabular}[c]{@{}l@{}}\rebuttal{This agent captures the phase competition}  \\ \rebuttal{relation between traffic movements through} \\ \rebuttal{a modified network structure.} \end{tabular}    
\\ \hline
\rebuttal{MPLight}        &    \rebuttal{pressure, phase}   & \rebuttal{Acylic}   &   \rebuttal{pressure}      & \rebuttal{Q-Learning}  & \begin{tabular}[c]{@{}l@{}} \rebuttal{This agent is based on FRAP and integrates} \\\rebuttal{pressure into state and reward design.}  \end{tabular}

\\ \hline
\end{tabular}}
\end{table}
\paragraph{Extensible design}
\ours provides a flexible interface to help user\rebuttal{s} customize their own RL model and RL design (state, reward, and action). Users can define their model through \texttt{Agent} module by completing abstract methods predefined in \texttt{BaseAgent} class. 
Existing RL libraries like \texttt{pfrl} can also be integrated into \texttt{Agent} class. \ours also provide different state and reward functions by instantiating \texttt{Generator} with subscribed function names in \texttt{info\_functions} to retrieve queue length, pressure, average lane speed, etc. 
Users could also customize their own reward or state functions by constructing a key-value mapping between new defined functions and \texttt{info\_functions}, which could be carried to \texttt{Agent} later by \texttt{Generator} class. 
\section{Experiment}
In this section, we first present our result and verify that our implementation is consistent with previous publications. 
In the second part, we compare different algorithms' performance with different datasets in both SUMO and CityFlow. 
Finally, we test the feasibility of \ours to verify it can properly run on large-scale and complex road net. Further, we also adapt algorithms from wide\rebuttal{ly} used RL library to testify our \texttt{Agent} module is flexible and easy to manipulate. Along the experiments, we will discuss the answers to several questions that motivate \ours: \emph{ Which simulator should I conduct experiments on?} \emph{Which evaluation metrics should I use?} \emph{Which RL method should I choose?} \emph{Is \ours suitable for my research?}

\subsection{Validation and Calibration}
\label{sec:exp:validation}
For testifying our PyTorch benchmark algorithms implementation, we compare the learning curve\rebuttal{s} and final performance\rebuttal{s} of the RL algorithms originally implemented in TensorFlow library. The simulator setting and observed traffic information are chosen to be similar to those used in  previous publications.

\begin{figure*}[tbh]
\centering
  \begin{tabular}{ccc}
  \includegraphics[width=.45\linewidth]{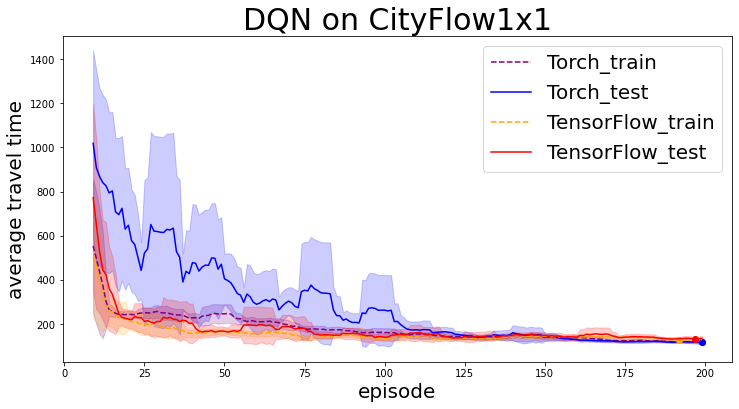} &
  \includegraphics[width=.45\linewidth]{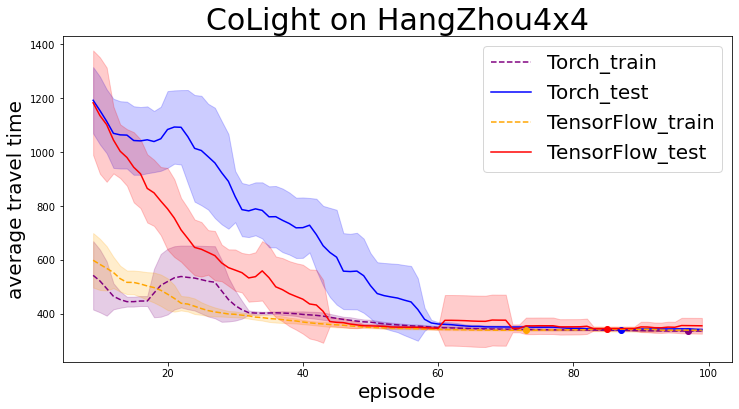} \\
  (a) \rebuttal{\dqn} in \gridone & (b) \rebuttal{\colight} in \hzfour  \\
  \end{tabular}
 \caption{Convergence curve of models implemented in their original form (Tensorflow) and in \ours (PyTorch). Y-axis are the testing results w.r.t. average travel time (in seconds). Validation for more models can be found in Appendix.}
    \label{fig:torch calibration}
\end{figure*}

\textbf{TensorFlow to PyTorch validation}
To validate the model's performance under the framework, \ours re-implemented some of the previous models from Tensorflow with PyTorch. For example, \rebuttal{\dqn}\rebuttal{~\cite{https://doi.org/10.48550/arxiv.1905.04716}} and \rebuttal{\colight}~\cite{wei2019colight} are originally \rebuttal{implemented} in Tensorflow, we reproduce the experiments of these models and  compare their performance in CityFlow simulator. 
Figure~\ref{fig:torch calibration} presents the learning curves and final performance of the original TensorFlow and our PyTorch implementation. On \gridone, the average travel time (in seconds) of our \rebuttal{\dqn} implementation converges to 116.28 which is also close to the original's 127.07 from\rebuttal{~\cite{https://doi.org/10.48550/arxiv.1905.04716}}. Our \rebuttal{\colight} implementation converges to 344.41 on average travel time metric, which is close to TensorFlow's 344.49 from~\cite{wei2019colight}.

\begin{figure*}[tbh]
\centering
  \includegraphics[width=.95\linewidth]{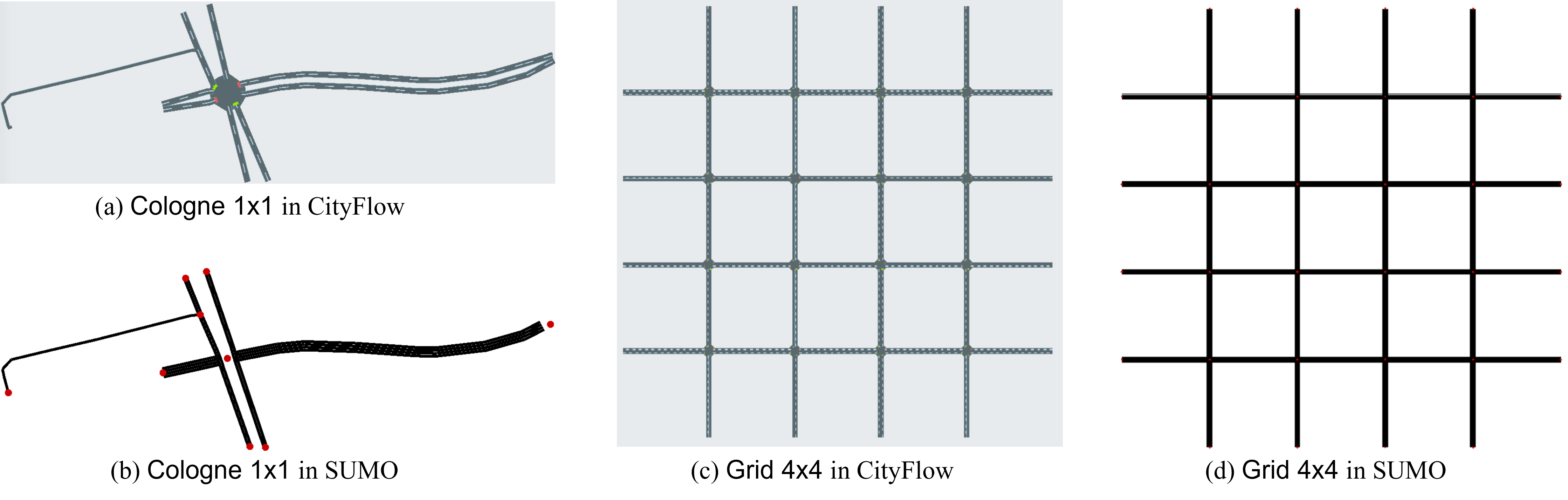} 
 \caption{Road networks in different simulators for calibration \rebuttal{The pictures with a gray background are the visualization of networks in the SUMO simulator, and the ones with the white background are in CityFlow. These are the outlines of traffic structures transferred between each other}(More networks can be found in Appendix).}
    \label{fig:tf-th}
\end{figure*}

\textbf{SUMO and CityFlow calibration}.
To validate the algorithms performance\rebuttal{s} are consistent in both SUMO and CityFlow, we calibrate under three road network\rebuttal{s} \gridfour, \realone and \hzfour. Their road network is shown in Figure~\ref{fig:tf-th}. In addition, we compare \rebuttal{\maxp}, \rebuttal{\sotl} and \rebuttal{\ft} algorithms performance since these three algorithms are deterministic given fixed network and traffic flow files.  Table~\ref{tab:calibration experiment} shows the overall performance before and after calibration. We have the following observations:

 \begin{table}[tbh]
\centering
\caption{Performance comparison of agents w.r.t. average travel time (in seconds) before and after calibration}
\label{tab:calibration experiment}
\resizebox{\textwidth}{!}{
\begin{tabular}{|c|cccccc|ccccll|}
\hline
\textbf{Calibration} &
  \multicolumn{6}{c|}{\textbf{Before}} &
  \multicolumn{6}{c|}{\textbf{After}} \\ \hline
\textbf{Dataset} &
  \multicolumn{2}{c|}{\textbf{Grid4x4}} &
  \multicolumn{2}{c|}{\textbf{Cologne1x1}} &
  \multicolumn{2}{c|}{\textbf{HangZhou4x4}} &
  \multicolumn{2}{c|}{\textbf{Grid4x4}} &
  \multicolumn{2}{c|}{\textbf{Cologne1x1}} &
  \multicolumn{2}{c|}{\textbf{HangZhou4x4}} \\ \hline
\textbf{Avg. Travel Time} &
  \multicolumn{1}{c|}{\textbf{Cityflow}} &
  \multicolumn{1}{c|}{\textbf{SUMO}} &
  \multicolumn{1}{c|}{\textbf{Cityflow}} &
  \multicolumn{1}{c|}{\textbf{SUMO}} &
  \multicolumn{1}{c|}{\textbf{Cityflow}} &
  \textbf{SUMO} &
  \multicolumn{1}{c|}{\textbf{Cityflow}} &
  \multicolumn{1}{c|}{\textbf{SUMO}} &
  \multicolumn{1}{c|}{\textbf{Cityflow}} &
  \multicolumn{1}{c|}{\textbf{SUMO}} &
  \multicolumn{1}{c|}{\textbf{Cityflow}} &
  \multicolumn{1}{c|}{\textbf{SUMO}} \\ \hline
\textbf{MaxPressure} &
  \multicolumn{1}{c|}{161.2878} &
  \multicolumn{1}{c|}{143.8296} &
  \multicolumn{1}{c|}{51.2785} &
  \multicolumn{1}{c|}{56.9068} &
  \multicolumn{1}{c|}{365.0634} &
  255.0052 &
  \multicolumn{1}{c|}{142.5275} &
  \multicolumn{1}{c|}{143.8296} &
  \multicolumn{1}{c|}{58.3441} &
  \multicolumn{1}{c|}{56.9068} &
  \multicolumn{1}{l|}{365.0634} &
  350.4125 \\ \hline
\textbf{FixedTime} &
  \multicolumn{1}{c|}{290.9525} &
  \multicolumn{1}{c|}{194.6542} &
  \multicolumn{1}{c|}{156.6599} &
  \multicolumn{1}{c|}{257.4194} &
  \multicolumn{1}{c|}{689.0221} &
  300.6685 &
  \multicolumn{1}{c|}{179.6606} &
  \multicolumn{1}{c|}{194.6542} &
  \multicolumn{1}{c|}{206.4620} &
  \multicolumn{1}{c|}{257.4194} &
  \multicolumn{1}{l|}{689.0221} &
  535.0060 \\ \hline
\textbf{SOTL} &
  \multicolumn{1}{c|}{185.8846} &
  \multicolumn{1}{c|}{208.3936} &
  \multicolumn{1}{c|}{1356.6037} &
  \multicolumn{1}{c|}{49.9770} &
  \multicolumn{1}{c|}{354.1250} &
  389.7881 &
  \multicolumn{1}{c|}{187.8568} &
  \multicolumn{1}{c|}{208.3936} &
  \multicolumn{1}{c|}{50.5813} &
  \multicolumn{1}{c|}{49.9770} &
  \multicolumn{1}{l|}{354.1250} &
  386.7881 \\ \hline
\end{tabular}
}
\end{table}

~\noindent\\~$\bullet$ Under gird-like networks (\gridfour, \hzfour), SUMO and CityFlow could achieve similar performance. Different agents' performance is not identical across simulators, but their rank within the same simulator is relatively consistent. 
~\noindent\\~$\bullet$ The discrepancy appears under more complex networks like \realone before calibration. Before calibration, we can see that \rebuttal{\ft} and \rebuttal{\sotl} perform worse than \rebuttal{\maxp}. After calibration, the same agent's performance is close and within an acceptable discrepancy between SUMO and CityFlow. Moreover, the ranking of the performances for different agents is consistent across different simulators after calibration. To double check our calibration is correct, \rebuttal{\dqn} algorithms are also trained under \realone with different simulators. In CityFlow and SUMO simulator, the result\rebuttal{s are} 50.581 and 49.977 in \realone w.r.t average travel time (in seconds), which further proved our calibration is correct.

\rebuttal{We modified some default settings in SUMO to compare the algorithms' performance under different simulators as comprehensively and fairly as possible, e.g., disabled the feature of dynamic routing and set teleport to be -1.} But it is worth noting that there are \rebuttal{still} some discrepanc\rebuttal{ies} that our current calibration cannot address. For example, SUMO has pedestrian traffic signals that CityFlow does not support, where vehicles will decelerate when they encounter the pedestrian traffic signals in SUMO and in CityFlow the vehicles will pass the intersection at normal speed. In addition, the vehicles in SUMO will also randomly decelerate when approaching a traffic signal even though the traffic signal is located further away; \rebuttal{I}n comparison, the vehicles in CityFlow will not have such randomness.

\subsection{Overall Performance}
\label{sec:exp:overall}
For comparative purposes, all benchmark RL models and traditional traffic control algorithms were compared under different simulator environments. All observations and rewards are set to be the same if not specifically mentioned, and all the hyperparameters are set according to the original implementations. We represent the final results truncated at 200 training iterations for a fair comparison, since most algorithms could converge within this period. While \rebuttal{\ppo} and \rebuttal{\maddpg} are noticeable for their high demand for training time, we provide the full converge curve in the Appendix. The results are \rebuttal{summarized} in Table~\ref{tab:comparison-experiment}. 
We have the following observations:
~\noindent\\~$\bullet$ Under the same dataset and simulator, the performance of the same model varies w.r.t. current four metrics. \emph{Travel time and queue length} are consistent with each other in most cases. \rebuttal{\emph{Throughput} sometimes is hard to differentiate the identical results under certain datasets.} For example, in \gridfour, all the methods served the same number of vehicles. \emph{Delay} sometime aligns with \emph{queue length} and \emph{travel time}, but can be different from all the other three metrics in some cases, e.g., \realthree under CityFlow and \gridfour under SUMO. This is because the delay is approximated from the average speed proposed by~\cite{ault2021reinforcement} and is not the actual delay calculated by vehicles' total travel time and desired travel time under maximum speed.
~\noindent\\~$\bullet$ Traditional transportation methods like \rebuttal{\maxp} can achieve consistent satisfactory performance though it is not the best. \rebuttal{\dqn} performs the best in single intersection scenarios. With more complicated road networks like~\realone and~\realthree, \rebuttal{\presslight} achieves better performance.


\begin{table}[tbh]
\centering
\caption{ Performance of agents in CityFlow and SUMO with \underline{\textbf{best}} and \underline{second best} performance highlighted.}
\label{tab:comparison-experiment}
\resizebox{\textwidth}{!}{
\begin{tabular}{|c|cccccccc|}
\hline
Network &
  \multicolumn{8}{c|}{Cityflow 1x1} \\ \hline
Simulator &
  \multicolumn{4}{c|}{CityFlow} &
  \multicolumn{4}{c|}{SUMO} \\ \hline
Metric &
  \multicolumn{1}{c|}{Travel Time} &
  \multicolumn{1}{c|}{Queue} &
  \multicolumn{1}{c|}{Delay} &
  \multicolumn{1}{c|}{Throughput} &
  \multicolumn{1}{c|}{Travel Time} &
  \multicolumn{1}{c|}{Queue} &
  \multicolumn{1}{c|}{Delay} &
  Throughput \\ \hline
FixedTime(t\_fixed=10) &
  \multicolumn{1}{c|}{923.6665} &
  \multicolumn{1}{c|}{114.4167} &
  \multicolumn{1}{c|}{4.8309} &
  \multicolumn{1}{c|}{1006} &
  \multicolumn{1}{c|}{533.0330} &
  \multicolumn{1}{c|}{91.9500} &
  \multicolumn{1}{c|}{4.3162} &
  848 \\ \hline
\rebuttal{FixedTime(t\_fixed=30)} &
  \multicolumn{1}{c|}{\rebuttal{552.7249}} &
  \multicolumn{1}{c|}{\rebuttal{100.3444}} &
  \multicolumn{1}{c|}{\rebuttal{5.1862}} &
  \multicolumn{1}{c|}{\rebuttal{1455}} &
  \multicolumn{1}{c|}{\rebuttal{270.6999}} &
  \multicolumn{1}{c|}{\rebuttal{80.4389}} &
  \multicolumn{1}{c|}{\rebuttal{5.6198}} &
  \rebuttal{1546} \\ \hline
MaxPressure &
  \multicolumn{1}{c|}{303.8070} &
  \multicolumn{1}{c|}{100.9444} &
  \multicolumn{1}{c|}{6.7686} &
  \multicolumn{1}{c|}{1717} &
  \multicolumn{1}{c|}{137.9254} &
  \multicolumn{1}{c|}{43.3417} &
  \multicolumn{1}{c|}{6.2455} &
  1930 \\ \hline
SOTL &
  \multicolumn{1}{c|}{212.2152} &
  \multicolumn{1}{c|}{69.1056} &
  \multicolumn{1}{c|}{5.4247} &
  \multicolumn{1}{c|}{1861} &
  \multicolumn{1}{c|}{154.7708} &
  \multicolumn{1}{c|}{50.5583} &
  \multicolumn{1}{c|}{5.1124} &
  1915 \\ \hline
IDQN &
  \multicolumn{1}{c|}{116.6373} &
  \multicolumn{1}{c|}{28.3706} &
  \multicolumn{1}{c|}{\underline{\textbf{0.6111}}} &
  \multicolumn{1}{c|}{1959} &
  \multicolumn{1}{c|}{\underline{ 91.7185}} &
  \multicolumn{1}{c|}{\underline{ 16.4861}} &
  \multicolumn{1}{c|}{\underline{ \textbf{0.5397}}} &
  \underline{ 1968} \\ \hline
MAPG &
  \multicolumn{1}{c|}{490.7145} &
  \multicolumn{1}{c|}{118.6139} &
  \multicolumn{1}{c|}{0.7392} &
  \multicolumn{1}{c|}{1514} &
  \multicolumn{1}{c|}{235.1756} &
  \multicolumn{1}{c|}{108.7583} &
  \multicolumn{1}{c|}{0.7089} &
  655 \\ \hline
IPPO &
  \multicolumn{1}{c|}{308.8728} &
  \multicolumn{1}{c|}{95.3806} &
  \multicolumn{1}{c|}{0.7578} &
  \multicolumn{1}{c|}{1736} &
  \multicolumn{1}{c|}{182.7708} &
  \multicolumn{1}{c|}{79.1167} &
  \multicolumn{1}{c|}{0.7103} &
  1492 \\ \hline
PressLight &
  \multicolumn{1}{c|}{\underline{105.6764}} &
  \multicolumn{1}{c|}{\underline{\textbf{23.2028}}} &
  \multicolumn{1}{c|}{0.5754} &
  \multicolumn{1}{c|}{\underline{1965}} &
  \multicolumn{1}{c|}{97.5526} &
  \multicolumn{1}{c|}{20.0306} &
  \multicolumn{1}{c|}{0.5523} &
  \underline{\textbf{1969}} \\ \hline
\rebuttal{FRAP} &
  \multicolumn{1}{c|}{\underline{\rebuttal{\textbf{104.2751}}}} &
  \multicolumn{1}{c|}{\underline{\rebuttal{23.2944}}} &
  \multicolumn{1}{c|}{\underline{\rebuttal{0.6225}}} &
  \multicolumn{1}{c|}{\underline{\rebuttal{\textbf{1979}}}} &
  \multicolumn{1}{c|}{\underline{\rebuttal{\textbf{89.9246}}}} &
  \multicolumn{1}{c|}{\underline{\rebuttal{\textbf{14.9000}}}} &
  \multicolumn{1}{c|}{\underline{\rebuttal{0.5426}}} &
  \rebuttal{1963}\\ \hline
Network &
  \multicolumn{8}{c|}{Cologne 1x1} \\ \hline
Simulator &
  \multicolumn{4}{c|}{CityFlow} &
  \multicolumn{4}{c|}{SUMO} \\ \hline
Metric &
  \multicolumn{1}{c|}{Travel Time} &
  \multicolumn{1}{c|}{Queue} &
  \multicolumn{1}{c|}{Delay} &
  \multicolumn{1}{c|}{Throughput} &
  \multicolumn{1}{c|}{Travel Time} &
  \multicolumn{1}{c|}{Queue} &
  \multicolumn{1}{c|}{Delay} &
  Throughput \\ \hline
FixedTime(t\_fixed=10) &
  \multicolumn{1}{c|}{206.4620} &
  \multicolumn{1}{c|}{51.5889} &
  \multicolumn{1}{c|}{4.0432} &
  \multicolumn{1}{c|}{1847} &
  \multicolumn{1}{c|}{257.4194} &
  \multicolumn{1}{c|}{86.1139} &
  \multicolumn{1}{c|}{6.2857} &
  1471 \\ \hline
\rebuttal{FixedTime(t\_fixed=30)} &
  \multicolumn{1}{c|}{\rebuttal{156.9632}} &
  \multicolumn{1}{c|}{\rebuttal{46.2278}} &
  \multicolumn{1}{c|}{\rebuttal{4.2243}} &
  \multicolumn{1}{c|}{\rebuttal{1910}} &
  \multicolumn{1}{c|}{\rebuttal{109.2111}} &
  \multicolumn{1}{c|}{\rebuttal{40.2056}} &
  \multicolumn{1}{c|}{\rebuttal{5.4129}} &
   \rebuttal{1947} \\ \hline
MaxPressure &
  \multicolumn{1}{c|}{58.3441} &
  \multicolumn{1}{c|}{7.6944} &
  \multicolumn{1}{c|}{2.7739} &
  \multicolumn{1}{c|}{\underline{ \textbf{2002}}} &
  \multicolumn{1}{c|}{56.9068} &
  \multicolumn{1}{c|}{10.4306} &
  \multicolumn{1}{c|}{3.4879} &
  1996 \\ \hline
SOTL &
  \multicolumn{1}{c|}{1358.6191} &
  \multicolumn{1}{c|}{104.6083} &
  \multicolumn{1}{c|}{5.9367} &
  \multicolumn{1}{c|}{515} &
  \multicolumn{1}{c|}{277.9961} &
  \multicolumn{1}{c|}{97.7722} &
  \multicolumn{1}{c|}{7.1510} &
  511 \\ \hline
IDQN &
  \multicolumn{1}{c|}{\underline{ 50.5813}} &
  \multicolumn{1}{c|}{5.3667} &
  \multicolumn{1}{c|}{0.3117} &
  \multicolumn{1}{c|}{2000} &
  \multicolumn{1}{c|}{\underline{ \textbf{49.9770}}} &
  \multicolumn{1}{c|}{\underline{ \textbf{7.1250}}} &
  \multicolumn{1}{c|}{\underline{ \textbf{0.3828}}} &
  \underline{ \textbf{1999}} \\ \hline
MAPG &
  \multicolumn{1}{c|}{\underline{ \textbf{46.7260}}} &
  \multicolumn{1}{c|}{\underline{ 5.0639}} &
  \multicolumn{1}{c|}{\underline{ 0.3002}} &
  \multicolumn{1}{c|}{2000} &
  \multicolumn{1}{c|}{1105.5855} &
  \multicolumn{1}{c|}{64.7528} &
  \multicolumn{1}{c|}{0.4953} &
  152 \\ \hline
IPPO &
  \multicolumn{1}{c|}{55.7449} &
  \multicolumn{1}{c|}{7.3500} &
  \multicolumn{1}{c|}{0.3334} &
  \multicolumn{1}{c|}{1994} &
  \multicolumn{1}{c|}{1134.2086} &
  \multicolumn{1}{c|}{64.4167} &
  \multicolumn{1}{c|}{0.4941} &
  163 \\ \hline
PressLight &
  \multicolumn{1}{c|}{51.7887} &
  \multicolumn{1}{c|}{\underline{ \textbf{4.8222}}} &
  \multicolumn{1}{c|}{\underline{ \textbf{0.2940}}} &
  \multicolumn{1}{c|}{\underline{ 2001}} &
  \multicolumn{1}{c|}{\underline{ 54.5398}} &
  \multicolumn{1}{c|}{\underline{ 9.4028}} &
  \multicolumn{1}{c|}{\underline{ 0.4199}} &
  \underline{ 1997} \\ \hline
\rebuttal{FRAP} &
  \multicolumn{1}{c|}{\rebuttal{66.6718}} &
  \multicolumn{1}{c|}{\rebuttal{12.0889}} &
  \multicolumn{1}{c|}{\rebuttal{0.3094}} &
  \multicolumn{1}{c|}{\rebuttal{1995}} &
  \multicolumn{1}{c|}{\rebuttal{93.4075}} &
  \multicolumn{1}{c|}{\rebuttal{29.3444}} &
  \multicolumn{1}{c|}{\rebuttal{0.5994}} &
  \rebuttal{1951}\\ \hline
Network &
  \multicolumn{8}{c|}{Cologne 1x3} \\ \hline
Simulator &
  \multicolumn{4}{c|}{CityFlow} &
  \multicolumn{4}{c|}{SUMO} \\ \hline
Metric &
  \multicolumn{1}{c|}{Travel Time} &
  \multicolumn{1}{c|}{Queue} &
  \multicolumn{1}{c|}{Delay} &
  \multicolumn{1}{c|}{Throughput} &
  \multicolumn{1}{c|}{Travel Time} &
  \multicolumn{1}{c|}{Queue} &
  \multicolumn{1}{c|}{Delay} &
  Throughput \\ \hline
FixedTime(t\_fixed=10) &
  \multicolumn{1}{c|}{93.3808} &
  \multicolumn{1}{c|}{4.8028} &
  \multicolumn{1}{c|}{1.5649} &
  \multicolumn{1}{c|}{2791} &
  \multicolumn{1}{c|}{108.0544} &
  \multicolumn{1}{c|}{10.0991} &
  \multicolumn{1}{c|}{2.5904} &
  2793 \\ \hline
\rebuttal{FixedTime(t\_fixed=30)} &
  \multicolumn{1}{c|}{\rebuttal{91.9379}} &
  \multicolumn{1}{c|}{\rebuttal{5.9120}} &
  \multicolumn{1}{c|}{\rebuttal{1.7413}} &
  \multicolumn{1}{c|}{\rebuttal{2781}} &
  \multicolumn{1}{c|}{\rebuttal{118.0659}} &
  \multicolumn{1}{c|}{\rebuttal{9.9722}} &
  \multicolumn{1}{c|}{\rebuttal{3.1455}} &
   \rebuttal{2794} \\ \hline
MaxPressure &
  \multicolumn{1}{c|}{56.6100} &
  \multicolumn{1}{c|}{1.5667} &
  \multicolumn{1}{c|}{1.1976} &
  \multicolumn{1}{c|}{2787} &
  \multicolumn{1}{c|}{\underline{ \textbf{59.5518}}} &
  \multicolumn{1}{c|}{\underline{ 1.2852}} &
  \multicolumn{1}{c|}{1.3130} &
  2818 \\ \hline
SOTL &
  \multicolumn{1}{c|}{1413.4136} &
  \multicolumn{1}{c|}{48.5324} &
  \multicolumn{1}{c|}{4.1193} &
  \multicolumn{1}{c|}{735} &
  \multicolumn{1}{c|}{306.5369} &
  \multicolumn{1}{c|}{53.8917} &
  \multicolumn{1}{c|}{5.1416} &
  1058 \\ \hline
IDQN &
  \multicolumn{1}{c|}{69.5980} &
  \multicolumn{1}{c|}{2.0667} &
  \multicolumn{1}{c|}{0.2076} &
  \multicolumn{1}{c|}{\underline{ \textbf{3509}}} &
  \multicolumn{1}{c|}{\underline{ 60.1986}} &
  \multicolumn{1}{c|}{\underline{ \textbf{1.1225}}} &
  \multicolumn{1}{c|}{\underline{ \textbf{0.1848}}} &
  2820 \\ \hline
MAPG &
  \multicolumn{1}{c|}{67.1616} &
  \multicolumn{1}{c|}{6.6310} &
  \multicolumn{1}{c|}{0.1769} &
  \multicolumn{1}{c|}{1204} &
  \multicolumn{1}{c|}{1276.6503} &
  \multicolumn{1}{c|}{49.4509} &
  \multicolumn{1}{c|}{0.6361} &
  1086 \\ \hline
IPPO &
  \multicolumn{1}{c|}{57.2221} &
  \multicolumn{1}{c|}{1.6370} &
  \multicolumn{1}{c|}{0.2679} &
  \multicolumn{1}{c|}{\underline{ 2792}} &
  \multicolumn{1}{c|}{70.0917} &
  \multicolumn{1}{c|}{3.2630} &
  \multicolumn{1}{c|}{0.2962} &
  2813 \\ \hline
PressLight &
  \multicolumn{1}{c|}{\underline{ 53.4214}} &
  \multicolumn{1}{c|}{\underline{ \textbf{0.8065}}} &
  \multicolumn{1}{c|}{\underline{ 0.1284}} &
  \multicolumn{1}{c|}{2790} &
  \multicolumn{1}{c|}{77.0971} &
  \multicolumn{1}{c|}{2.9648} &
  \multicolumn{1}{c|}{\underline{ 0.2139}} &
  \underline{\textbf{2822}} \\ \hline
\rebuttal{MPLight} &
  \multicolumn{1}{c|}{\underline{\rebuttal{\textbf{52.9214}}}} &
  \multicolumn{1}{c|}{\rebuttal{\underline{1.2926}}} &
  \multicolumn{1}{c|}{\underline{\rebuttal{\textbf{0.1271}}}} &
  \multicolumn{1}{c|}{\rebuttal{2790}} &
  \multicolumn{1}{c|}{\rebuttal{96.1173}} &
  \multicolumn{1}{c|}{\rebuttal{3.4630}} &
  \multicolumn{1}{c|}{\rebuttal{0.2282}} &
  \underline{\rebuttal{2821}}\\ \hline
Network &
  \multicolumn{8}{c|}{Grid4x4} \\ \hline
Simulator &
  \multicolumn{4}{c|}{CityFlow} &
  \multicolumn{4}{c|}{SUMO} \\ \hline
Metric &
  \multicolumn{1}{c|}{Travel Time} &
  \multicolumn{1}{c|}{Queue} &
  \multicolumn{1}{c|}{Delay} &
  \multicolumn{1}{c|}{Throughput} &
  \multicolumn{1}{c|}{Travel Time} &
  \multicolumn{1}{c|}{Queue} &
  \multicolumn{1}{c|}{Delay} &
  Throughput \\ \hline
FixedTime(t\_fixed=10) &
  \multicolumn{1}{c|}{179.6606} &
  \multicolumn{1}{c|}{1.2889} &
  \multicolumn{1}{c|}{1.2128} &
  \multicolumn{1}{c|}{\underline{\textbf{1473}}} &
  \multicolumn{1}{c|}{194.6542} &
  \multicolumn{1}{c|}{1.9967} &
  \multicolumn{1}{c|}{1.2571} &
  1443 \\ \hline
\rebuttal{FixedTime(t\_fixed=30)} &
  \multicolumn{1}{c|}{\rebuttal{258.1894}} &
  \multicolumn{1}{c|}{\rebuttal{3.3891}} &
  \multicolumn{1}{c|}{\rebuttal{2.0448}} &
  \multicolumn{1}{c|}{\rebuttal{1465}} &
  \multicolumn{1}{c|}{\rebuttal{292.0315}} &
  \multicolumn{1}{c|}{\rebuttal{4.4988}} &
  \multicolumn{1}{c|}{\rebuttal{2.2300}} &
   \rebuttal{1428} \\ \hline
MaxPressure &
  \multicolumn{1}{c|}{142.8568} &
  \multicolumn{1}{c|}{0.5010} &
  \multicolumn{1}{c|}{0.6866} &
  \multicolumn{1}{c|}{\underline{\textbf{1473}}} &
  \multicolumn{1}{c|}{\underline{ 143.8296}} &
  \multicolumn{1}{c|}{\underline{ 0.6276}} &
  \multicolumn{1}{c|}{0.5197} &
  1461 \\ \hline
SOTL &
  \multicolumn{1}{c|}{187.8568} &
  \multicolumn{1}{c|}{1.5012} &
  \multicolumn{1}{c|}{1.2580} &
  \multicolumn{1}{c|}{\underline{\textbf{1473}}} &
  \multicolumn{1}{c|}{208.3936} &
  \multicolumn{1}{c|}{2.3321} &
  \multicolumn{1}{c|}{1.3739} &
  1443 \\ \hline
IDQN &
  \multicolumn{1}{c|}{\underline{\textbf{133.8038}}} &
  \multicolumn{1}{c|}{\underline{\textbf{0.2540}}} &
  \multicolumn{1}{c|}{\underline{\textbf{0.0411}}} &
  \multicolumn{1}{c|}{\underline{\textbf{1473}}} &
  \multicolumn{1}{c|}{\underline{\textbf{143.3112}}} &
  \multicolumn{1}{c|}{\underline{\textbf{0.5997}}} &
  \multicolumn{1}{c|}{\underline{\textbf{0.0402}}} &
  \underline{1462} \\ \hline
MAPG &
  \multicolumn{1}{c|}{227.6490} &
  \multicolumn{1}{c|}{2.4816} &
  \multicolumn{1}{c|}{0.1259} &
  \multicolumn{1}{c|}{\underline{\textbf{1473}}} &
  \multicolumn{1}{c|}{225.4148} &
  \multicolumn{1}{c|}{2.4328} &
  \multicolumn{1}{c|}{0.1277} &
  \underline{\textbf{1471}}
   \\ \hline
IPPO &
  \multicolumn{1}{c|}{197.8900} &
  \multicolumn{1}{c|}{1.7667} &
  \multicolumn{1}{c|}{0.1134} &
  \multicolumn{1}{c|}{\underline{\textbf{1473}}} &
  \multicolumn{1}{c|}{206.2505} &
  \multicolumn{1}{c|}{2.3151} &
  \multicolumn{1}{c|}{0.1154} &
  1437 \\ \hline
PressLight &
  \multicolumn{1}{c|}{142.3299} &
  \multicolumn{1}{c|}{0.4528} &
  \multicolumn{1}{c|}{0.0542} &
  \multicolumn{1}{c|}{\underline{\textbf{1473}}} &
  \multicolumn{1}{c|}{147.2203} &
  \multicolumn{1}{c|}{0.6978} &
  \multicolumn{1}{c|}{0.0456} &
  \underline{1462} \\ \hline
CoLight &
  \multicolumn{1}{c|}{\underline{ 138.6857}} &
  \multicolumn{1}{c|}{0.4038} &
  \multicolumn{1}{c|}{\underline{ 0.0476}} &
  \multicolumn{1}{c|}{\underline{\textbf{1473}}} &
  \multicolumn{1}{c|}{143.9758} &
  \multicolumn{1}{c|}{0.6288} &
  \multicolumn{1}{c|}{\underline{0.0437}} &
  1461 \\ \hline
\rebuttal{MPLight} &
  \multicolumn{1}{c|}{\rebuttal{140.8452}}&
  \multicolumn{1}{c|}{\underline{\rebuttal{0.3976}}} &
  \multicolumn{1}{c|}{\rebuttal{0.0513}} &
  \multicolumn{1}{c|}{\underline{\rebuttal{\textbf{1473}}}} &
  \multicolumn{1}{c|}{\rebuttal{158.4627}} &
  \multicolumn{1}{c|}{\rebuttal{1.0156}} &
  \multicolumn{1}{c|}{\rebuttal{0.0600}} &
  \rebuttal{1461}\\ \hline
\end{tabular}
}
\end{table}

We also conducted experiments for different network scalability and complexity, and the results can be found in the Appendix.

\subsection{Discussion}
\label{sec:discussion}

\paragraph{Which simulator should I conduct experiments on?}
From the running time comparison in Table~\ref{tab:Running Time} between SUMO and CityFlow simulator, we find that  CityFlow \rebuttal{and SUMO (with Libsumo)}'s time cost is around ten times less than SUMO \rebuttal{(with TraCI)} which indicates its higher running efficiency. \rebuttal{Different from CityFlow,} SUMO provides a more accurate depiction of vehicles' state and more complex traffic operations, including changing lanes and 'U-turn'\rebuttal{. Also, SUMO provides users with more realistic settings, including pedestrians, driver imperfection, collisions, and dynamic routing.} Thus, it is more powerful on complex networks \rebuttal{ and reflecting real-\rebuttal{world} scenario}.
\begin{table}[htb]
\centering
\caption{Performance comparison w.r.t. the running time (in seconds) of different methods under two simulators.}
\label{tab:Running Time}
\resizebox{\textwidth}{!}{
\begin{tabular}{|c|c|c|c|c|c|c|c|}
\hline
\textbf{Running Time} & \textbf{Simulator} & \textbf{FixedTime} & \textbf{MaxPressure} & \textbf{SOTL} & \textbf{IDQN} & \textbf{IPPO} & \textbf{PressLight} \\ \hline
\multirow{2}{*}{\textbf{Cityflow1x1}} & \textbf{CityFlow} & 5.7593   & 9.5857  & 6.0272  & 2461.7496  & 2450.7397  & 1960.7932  \\ \cline{2-8} 
                                      & \rebuttal{\textbf{SUMO(Libsumo)}}     & \rebuttal{6.0006}  & \rebuttal{4.2174} & \rebuttal{4.2988}  & \rebuttal{3691.8403} & \rebuttal{2833.0523} & \rebuttal{3619.1282} \\ \cline{2-8} 
                                      & \rebuttal{\textbf{SUMO(Traci)}}     & 73.6791  & 46.5712 & 51.488  & 30279.5201 & 37662.9677 & 27697.9098 \\ \hline
\multirow{2}{*}{\textbf{Cologne1x1}}  & \textbf{CityFlow} & 3.3049   & 2.7601  & 4.6527  & 1641.6128  & 3148.0649  & 3066.7189  \\ \cline{2-8} 
                                        & \rebuttal{\textbf{SUMO(Libsumo)}}     & \rebuttal{4.3649} & \rebuttal{3.1411} & \rebuttal{3.2771} & \rebuttal{2649.0341} & \rebuttal{2581.8182} & \rebuttal{2863.9526} \\ \cline{2-8}
                                      & \rebuttal{\textbf{SUMO(Traci)}}     & 133.6334 & 24.9327 & 71.7565 & 11243.2917 & 31431.1760 & 11535.0851 \\ \hline 
                                      
\end{tabular}
}
\end{table}

\paragraph{Which evaluation metrics should I use?}
Average travel time is generally a good metric to evaluate algorithms' performance on traffic control tasks. \rebuttal{But for settings with dynamic routing, the travel time would not be a good metric as the average travel time of a vehicle can change with dynamic routing. In \ours, the simulation under SUMO disabled the feature of dynamic routing so the travel time would be good on the current settings.} From Table~\ref{tab:comparison-experiment}, we can see that lane delay and throughput are not always consistent between different simulators and even in \rebuttal{the} same simulator environment. Sometimes they often show contradictory performances in different datasets. Therefore, we suggest researcher report travel time as a necessity, and other metrics of their interest in their papers.

\paragraph{Which RL method should I choose?}
From our experiments in Table~\ref{tab:comparison-experiment}, we can see that Actor-Critic based RL algorithms need a long time to converge. In \realone, \rebuttal{\maddpg} and \rebuttal{\ppo} algorithms still perform badly after 200 iterations. \rebuttal{\dqn}  and other Q-learning-based algorithms are generally good choices in all five datasets. We can see that they outperform traditional non-RL algorithms all the time. Comparing results in \gridone and \realthree, we find \rebuttal{\frap} and \rebuttal{\mplight} could bring improvement compared to \rebuttal{\dqn} algorithm.

\paragraph{When should I use \ours?}
Since \ours provides a highly unified interface to help users choose or define their functions \rebuttal{and} extract information from the simulator's environment, it is a powerful platform for users to investigate the best combination of state and reward functions for current state-of-the-art or their implemented models. Also, users could compare their algorithms with our implemented baseline model using the evaluation metrics we provided. In addition, since \ours supports multiple simulation environments\rebuttal{,} users could also conduct experiments in the different simulation environments to validate \rebuttal{that }their algorithms \rebuttal{are} robust and achieve generally good performance under different settings.
\paragraph{When shouldn't I use \ours?}
Currently, \ours only supports SUMO and CityFlow, thus users currently cannot run their experiments with other simulation engines in our library unless implementing an \texttt{World} like \ours did for SUMO and CityFlow. Also, users might need to spend extensive labor to compare their experiment result\rebuttal{s} across simulators if they use their datasets because of dataset calibration. Furthermore, if users plan to use algorithms with extra information like \rebuttal{\frap}~\cite{zheng2019learning}\rebuttal{,} they might need to define competitive phases, and the implementation of their agent\rebuttal{s} under complex or large-scale network \rebuttal{which} should be rather complicated. 
\section{Conclusion}
In this paper, we introduced \ours, a highly unified, extensible, and comprehensive library for traffic light control tasks. We collected and filtered nine commonly used datasets and implemented ten different baseline models across two influential traffic simulators, including SUMO and CityFlow. We both conducted experiments to prove our PyTorch implementation could achieve the same level of performance as the original TensorFlow official code and calibrated simulators to improve the reliability of our cross-simulator environment. Moreover, the performance of all implemented algorithms \rebuttal{was} compared under various datasets and simulators. We further provided the discussion for researchers interested in this topic with our benchmarking results.
In the future, we will implement more state-of-the-art RL-based algorithms and continu\rebuttal{ally} support more simulator environments. Further calibration efforts will be made to help different algorithms' performance comparisons across different simulators.  

\newpage




\newpage
\appendix

\section{Appendix}

\subsection{Documentation and License}
\ours is open source and free to use/modify under the GNU General Public License 3. The code and documents are available on Github at~\url{https://darl-libsignal.github.io/}.
The embedded traffic datasets are distributed with their own licenses from~\cite{websiteTSC} and~\cite{ault2021reinforcement}, whose licenses are under the GNU General Public License 3. SUMO is licensed under the EPL 2.0 and CityFlow is under Apache 2.0.  
All experiments can be reproduced from the source code, which includes all hyper-parameters and configuration.
The authors will bear all responsibility in case of violation of rights, etc., ensure access to the data and provide the necessary maintenance.

\subsection{Metrics Definition}

\textbf{Average travel time (travel time)}:
The average time that each vehicle spent on traveling within the network, including waiting time and actual travel time. A smaller travel time means the better performance.

\textbf{Queue length (queue)}:
The average queue length over time, where the queue length at time $t$ is the sum of the number of vehicles waiting on lanes. A smaller queue length means the better performance.

\textbf{Approximated delay (delay)}:
Averaged difference between the current speed of vehicle and the maximum speed limit of this lane over all vehicles, calculated from $ 1 - \frac{\sum_{i=1}^{n}v_i}{n * v_{max}}$~\cite{ault2021reinforcement}, where $n$ is the number of vehicles on the lane, $v_i$ is the speed of vehicle $i$ and $v_{max}$ is the maximum allowed speed. A smaller delay means the better performance.

\textbf{Real delay (real delay)}:
Real delay of a vehicle is defined as the time a vehicle has traveled within the environment minus the expected travel time. A smaller delay means better performance.

\textbf{Throughput}:
Number of vehicles that have finished their trips until current simulation step. A larger throughput means the better performance.

\subsection{Validation}
\begin{figure*}[tbh]
\centering
  \begin{tabular}{ccc}
  \includegraphics[width=.45\linewidth]{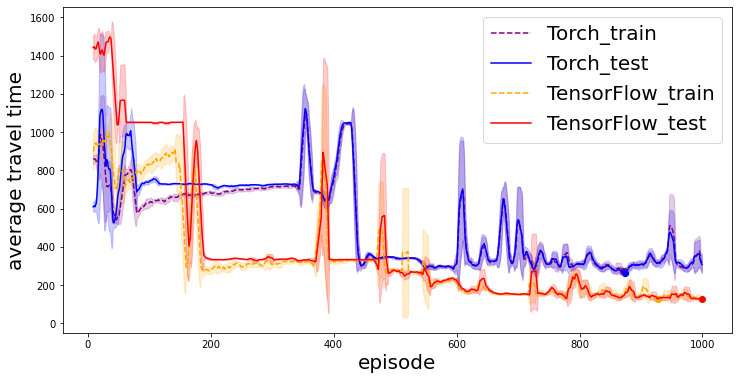} &
  \includegraphics[width=.45\linewidth]{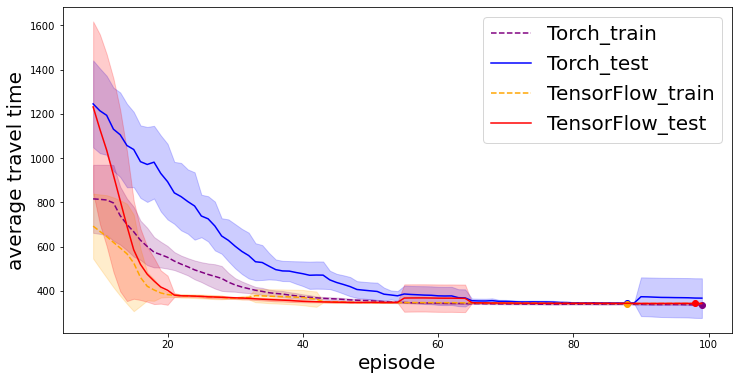} \\
  (a) \rebuttal{\maddpg} in \gridone & (b) \rebuttal{\presslight} in \hzfour  \\
  \includegraphics[width=.45\linewidth]{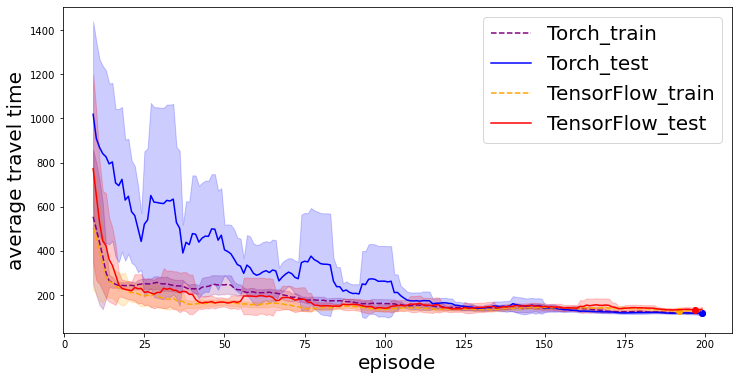} &
  \includegraphics[width=.45\linewidth]{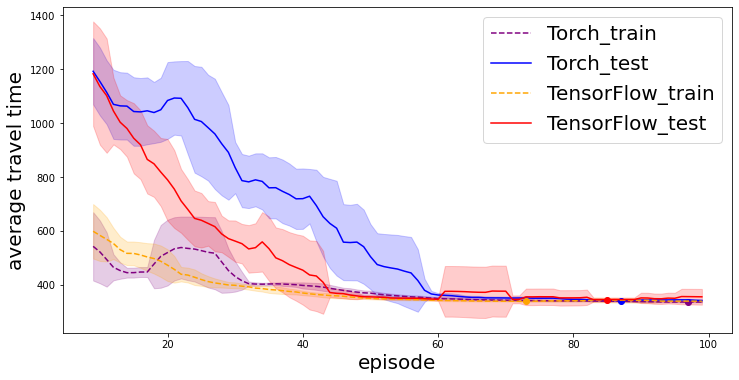} \\
  (a) \rebuttal{\dqn} in \gridone & (b) \rebuttal{\colight} in \hzfour  \\
  \end{tabular}
 \caption{Convergence curve of models implemented in their original form (TensorFlow) and in \ours (PyTorch). Y-axis are the testing results w.r.t. average travel time (in seconds).}
    \label{fig:torch calibration others}
\end{figure*}

To validate our PyTorch re-implementations performance, we compare the performance of four algorithms which originally implemented in TensorFlow. Fig~\ref{fig:torch calibration others} shows the converge curve of \rebuttal{\maddpg}, \rebuttal{\presslight}, \rebuttal{\dqn}, and \rebuttal{\colight} in both the train and test phase, which are not provided in Section~\ref{sec:exp:validation}. The final performance in Table~\ref{tab:validation sup} shows that all four new implementations are consistent with their original TensorFlow implementations. 

\begin{table}[tbh]
\centering
\caption{Best episode performance w.r.t. average travel time (in seconds). The performance of models are consistent under TensorFlow and PyTorch.}
\label{tab:validation sup}
\scalebox{1}{
\begin{tabular}{|c|c|c|}
\hline
\textbf{library}             & \textbf{TensorFlow} & \textbf{PyTorch} \\ \hline
\textbf{MAPG on CityFlow1x1}       & 125.786             & 180.608    \\ \hline
\textbf{IDQN on CityFlow1x1}        & 131.8               & 116.28           \\ \hline
\textbf{PressLight on HangZhou4x4} & 342.361             & 344.75           \\ \hline
\textbf{CoLight on HangZhou4x4}    & 344.49              & 341.41           \\ \hline
\end{tabular}
}
\end{table}

\subsection{Network Conversion}
\begin{figure*}[tbh]
\centering
  \includegraphics[width=1\linewidth]{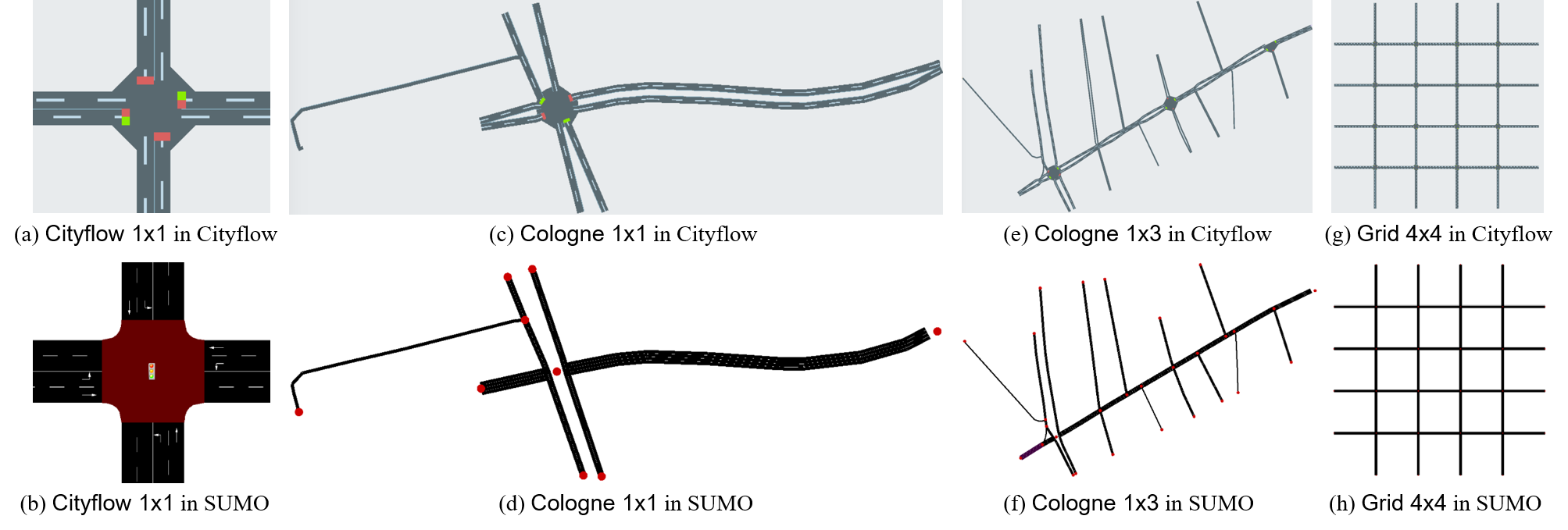} 
 \caption{Road networks in different simulators for calibration}
    \label{fig:roadvis}
\end{figure*}

Current \ours includes 9 datasets which are converted and calibrated. Their road networks are shown in Figure~\ref{fig:roadvis}. Other configuration of \rebuttal{\gridone} datasets are similar to \rebuttal{\gridone} appeared in full paper in road network structure, which will not be shown here.



\subsection{Calibration Steps}
To validate the performance of the algorithms are consistent in both SUMO and CityFlow, we calibrate the simulators in the following aspects: 
~\noindent\\$\bullet$~\textbf{Calibration from SUMO to CityFlow}: To make the conversion of complex networks from SUMO compatible with CityFlow, we redesign the original convert files from ~\cite{zhang2019cityflow} with the following: (1) For those \texttt{.rou} files in SUMO that only specify source and destination intersections and ignore roads that would be passing, the router command line in SUMO should be applied to generate full routes before converting it into CityFlow's \texttt{.json} traffic flow file. (2) We treat all the intersections without traffic signals in SUMO as ``virtual'' nodes in CityFlow's \texttt{.json} road network file. (3) We keep the time interval the same for red and yellow signals in SUMO and CityFlow. \rebuttal{(4) SUMO has a feature of the dynamic routing of vehicles that CityFlow does not have, currently all the simulation under SUMO in \ours disables the dynamic routing.}
~\noindent\\$\bullet$~\textbf{Calibration from CityFlow to SUMO}: The vehicles in CityFlow's traffic flow file need to be sorted according to their departure time because the SUMO traffic file defaults to the depart time of the preceding vehicle earlier than the following vehicle. 

\begin{table}[tbh]
\centering
\caption{Performance of agents in CityFlow and SUMO on additional datasets that are not shown in Section~\ref{sec:exp:overall} with \underline{\textbf{best}} and \underline{second best} performance highlighted.}
\label{tab:sup-comparison-exp}
\resizebox{\textwidth}{!}{
\begin{tabular}{|c|cccccccc|}
\hline
Network      & \multicolumn{8}{c|}{Cityflow 1x1(Config2)}                                                                                                                                                                                                                                                                                                         \\ \hline
Simulator    & \multicolumn{4}{c|}{CityFlow}                                                                                                                                                      & \multicolumn{4}{c|}{SUMO}                                                                                                                                     \\ \hline
Metric       & \multicolumn{1}{c|}{Travel Time}             & \multicolumn{1}{c|}{Queue}                  & \multicolumn{1}{c|}{Delay}                 & \multicolumn{1}{c|}{Throughput}          & \multicolumn{1}{c|}{Travel Time}             & \multicolumn{1}{c|}{Queue}                  & \multicolumn{1}{c|}{Delay}                 & Throughput          \\ \hline
FixedTime(t\_fixed=10)      & \multicolumn{1}{c|}{702.0847}                & \multicolumn{1}{c|}{90.3417}                & \multicolumn{1}{c|}{4.3542}                & \multicolumn{1}{c|}{914}                 & \multicolumn{1}{c|}{444.5625}                & \multicolumn{1}{c|}{74.9}                   & \multicolumn{1}{c|}{4.095}                 & 832                 \\ \hline
FixedTime(t\_fixed=30)      & \multicolumn{1}{c|}{\rebuttal{305.9428}}                & \multicolumn{1}{c|}{\rebuttal{62.9361}}                & \multicolumn{1}{c|}{\rebuttal{4.3038}}                & \multicolumn{1}{c|}{\rebuttal{1246}}                 & \multicolumn{1}{c|}{\rebuttal{181.3847}}                & \multicolumn{1}{c|}{\rebuttal{42.9972}}                   & \multicolumn{1}{c|}{\rebuttal{4.2947}}                 &      \rebuttal{1310}            \\ \hline
MaxPressurre & \multicolumn{1}{c|}{91.4333}                 & \multicolumn{1}{c|}{11.3806}                & \multicolumn{1}{c|}{4.5774}                & \multicolumn{1}{c|}{{\underline{1388}}}          & \multicolumn{1}{c|}{{\underline{ 85.7407}}}           & \multicolumn{1}{c|}{{\underline{ 8.7333}}}           & \multicolumn{1}{c|}{4.1472}                & {1377}          \\ \hline
SOTL         & \multicolumn{1}{c|}{116.2653}                & \multicolumn{1}{c|}{20.0139}                & \multicolumn{1}{c|}{3.6794}                & \multicolumn{1}{c|}{1381}                & \multicolumn{1}{c|}{96.8388}                 & \multicolumn{1}{c|}{13.2667}                & \multicolumn{1}{c|}{3.1959}                & 1371                \\ \hline
IDQN         & \multicolumn{1}{c|}{{\underline{ \textbf{79.6013}}}}  & \multicolumn{1}{c|}{{\underline{ \textbf{7.5806}}}}  & \multicolumn{1}{c|}{{\underline{ \textbf{0.4253}}}} & \multicolumn{1}{c|}{{\underline{ \textbf{1390}}}} & \multicolumn{1}{c|}{{\underline{ \textbf{78.0051}}}}  & \multicolumn{1}{c|}{{\underline{ \textbf{5.8111}}}}  & \multicolumn{1}{c|}{{\underline{ \textbf{0.3648}}}} & {\underline{ \textbf{1379}}} \\ \hline
MAPG         & \multicolumn{1}{c|}{262.1785}                & \multicolumn{1}{c|}{75.3056}                & \multicolumn{1}{c|}{0.6467}                & \multicolumn{1}{c|}{1237}                & \multicolumn{1}{c|}{125.3895}                & \multicolumn{1}{c|}{61.075}                 & \multicolumn{1}{c|}{0.5812}                & 955                 \\ \hline
IPPO         & \multicolumn{1}{c|}{733.7248}                & \multicolumn{1}{c|}{130.7694}               & \multicolumn{1}{c|}{0.6963}                & \multicolumn{1}{c|}{748}                 & \multicolumn{1}{c|}{90.9229}                 & \multicolumn{1}{c|}{10.6}                   & \multicolumn{1}{c|}{0.4447}                & 1375                \\ \hline
PressLight   & \multicolumn{1}{c|}{{\underline{ 84.7516}}}           & \multicolumn{1}{c|}{{\underline{ 9.2889}}}           & \multicolumn{1}{c|}{{\underline{ 0.4454}}}          & \multicolumn{1}{c|}{1385}                & \multicolumn{1}{c|}{85.873}                  & \multicolumn{1}{c|}{8.9639}                 & \multicolumn{1}{c|}{{\underline{ 0.4318}}}          & \underline{1378}                \\ \hline
Network      & \multicolumn{8}{c|}{Cityflow 1x1(Config3)}                                                                                                                                                                                                                                                                                                         \\ \hline
Simulator    & \multicolumn{4}{c|}{CityFlow}                                                                                                                                                      & \multicolumn{4}{c|}{SUMO}                                                                                                                                     \\ \hline
Metric       & \multicolumn{1}{c|}{Travel Time}             & \multicolumn{1}{c|}{Queue}                  & \multicolumn{1}{c|}{Delay}                 & \multicolumn{1}{c|}{Throughput}          & \multicolumn{1}{c|}{Travel Time}             & \multicolumn{1}{c|}{Queue}                  & \multicolumn{1}{c|}{Delay}                 & Throughput          \\ \hline
FixedTime(t\_fixed=10)      & \multicolumn{1}{c|}{461.2692}                & \multicolumn{1}{c|}{33.5944}                & \multicolumn{1}{c|}{2.3989}                & \multicolumn{1}{c|}{531}                 & \multicolumn{1}{c|}{284.4235}                & \multicolumn{1}{c|}{29.9806}                & \multicolumn{1}{c|}{2.2208}                & 503                 \\ \hline
FixedTime(t\_fixed=30)      & \multicolumn{1}{c|}{\rebuttal{228.8520}}                & \multicolumn{1}{c|}{\rebuttal{24.0667}}                & \multicolumn{1}{c|}{\rebuttal{2.7021}}                & \multicolumn{1}{c|}{\rebuttal{659}}                 & \multicolumn{1}{c|}{\rebuttal{173.1985}}                & \multicolumn{1}{c|}{\rebuttal{21.3167}}                & \multicolumn{1}{c|}{\rebuttal{2.8096}}                &  \rebuttal{680}                \\ \hline
MaxPressurre & \multicolumn{1}{c|}{69.7295}                 & \multicolumn{1}{c|}{2.2250}                 & \multicolumn{1}{c|}{1.8415}                & \multicolumn{1}{c|}{729}                 & \multicolumn{1}{c|}{69.9601}                 & \multicolumn{1}{c|}{{\underline{ 1.6167}}}           & \multicolumn{1}{c|}{1.4375}                & {\underline{ \textbf{726}}}  \\ \hline
SOTL         & \multicolumn{1}{c|}{89.2005}                 & \multicolumn{1}{c|}{5.5556}                 & \multicolumn{1}{c|}{1.8778}                & \multicolumn{1}{c|}{729}                 & \multicolumn{1}{c|}{84.9169}                 & \multicolumn{1}{c|}{4.4556}                 & \multicolumn{1}{c|}{1.7441}                & 722                 \\ \hline
IDQN         & \multicolumn{1}{c|}{{\underline{ \textbf{66.9865}}}}  & \multicolumn{1}{c|}{{\underline{ \textbf{1.6833}}}}  & \multicolumn{1}{c|}{{\underline{ \textbf{0.1816}}}} & \multicolumn{1}{c|}{{\underline{ 730}}}           & \multicolumn{1}{c|}{{\underline{ \textbf{68.7369}}}}  & \multicolumn{1}{c|}{{\underline{ \textbf{1.2556}}}}  & \multicolumn{1}{c|}{{\underline{ \textbf{0.1396}}}} & {\underline{ \textbf{726}}}  \\ \hline
MAPG         & \multicolumn{1}{c|}{183.2719}                & \multicolumn{1}{c|}{25.7778}                & \multicolumn{1}{c|}{0.4086}                & \multicolumn{1}{c|}{651}                 & \multicolumn{1}{c|}{115.5654}                & \multicolumn{1}{c|}{26.3361}                & \multicolumn{1}{c|}{0.3421}                & 543                 \\ \hline
IPPO         & \multicolumn{1}{c|}{79.4778}                 & \multicolumn{1}{c|}{3.8333}                 & \multicolumn{1}{c|}{0.2672}                & \multicolumn{1}{c|}{729}                 & \multicolumn{1}{c|}{78.8072}                 & \multicolumn{1}{c|}{3.0500}                 & \multicolumn{1}{c|}{0.2373}                & {\underline{ \textbf{726}}}  \\ \hline
PressLight   & \multicolumn{1}{c|}{{\underline{ 67.4899}}}           & \multicolumn{1}{c|}{{\underline{ 1.8861}}}           & \multicolumn{1}{c|}{{\underline{ 0.1945}}}          & \multicolumn{1}{c|}{{\underline{ \textbf{731}}}}  & \multicolumn{1}{c|}{73.4509}                 & \multicolumn{1}{c|}{2.4250}                 & \multicolumn{1}{c|}{{\underline{ 0.2169}}}          & {\underline{ 723}}           \\ \hline
Network      & \multicolumn{8}{c|}{Cityflow 1x1(Config4)}                                                                                                                                                                                                                                                                                                         \\ \hline
Simulator    & \multicolumn{4}{c|}{CityFlow}                                                                                                                                                      & \multicolumn{4}{c|}{SUMO}                                                                                                                                     \\ \hline
Metric       & \multicolumn{1}{c|}{Travel Time}             & \multicolumn{1}{c|}{Queue}                  & \multicolumn{1}{c|}{Delay}                 & \multicolumn{1}{c|}{Throughput}          & \multicolumn{1}{c|}{Travel Time}             & \multicolumn{1}{c|}{Queue}                  & \multicolumn{1}{c|}{Delay}                 & Throughput          \\ \hline
FixedTime(t\_fixed=10)      & \multicolumn{1}{c|}{686.7469}                & \multicolumn{1}{c|}{96.7222}                & \multicolumn{1}{c|}{4.4683}                & \multicolumn{1}{c|}{925}                 & \multicolumn{1}{c|}{469.5721}                & \multicolumn{1}{c|}{81.7778}                & \multicolumn{1}{c|}{4.0367}                & 811                 \\ \hline
FixedTime(t\_fixed=30)      & \multicolumn{1}{c|}{\rebuttal{339.3052}}                & \multicolumn{1}{c|}{\rebuttal{63.7750}}                & \multicolumn{1}{c|}{\rebuttal{4.4226}}                & \multicolumn{1}{c|}{\rebuttal{1290}}                 & \multicolumn{1}{c|}{\rebuttal{204.3520}}                & \multicolumn{1}{c|}{\rebuttal{51.7611}}                & \multicolumn{1}{c|}{\rebuttal{4.5938}}                &  \rebuttal{1358}                \\ \hline
MaxPressurre & \multicolumn{1}{c|}{136.5470}                & \multicolumn{1}{c|}{31.9083}                & \multicolumn{1}{c|}{5.1600}                & \multicolumn{1}{c|}{1556}                & \multicolumn{1}{c|}{98.8589}                 & \multicolumn{1}{c|}{17.3444}                & \multicolumn{1}{c|}{4.6113}                & 1602                \\ \hline
SOTL         & \multicolumn{1}{c|}{158.4722}                & \multicolumn{1}{c|}{40.2611}                & \multicolumn{1}{c|}{{\underline{ 4.5164}}}          & \multicolumn{1}{c|}{1562}                & \multicolumn{1}{c|}{113.0794}                & \multicolumn{1}{c|}{23.7389}                & \multicolumn{1}{c|}{3.7165}                & 1587                \\ \hline
IDQN         & \multicolumn{1}{c|}{{\underline{ 101.9282}} }         & \multicolumn{1}{c|}{{\underline{ 18.7278}} }         & \multicolumn{1}{c|}{0.5685}                & \multicolumn{1}{c|}{{\underline{ 1614}}}          & \multicolumn{1}{c|}{{\underline{ \textbf{90.0737}}}}  & \multicolumn{1}{c|}{{\underline{ \textbf{12.8278}}}} & \multicolumn{1}{c|}{{\underline{ 0.4912}}}          & {\underline{ \textbf{1614}}} \\ \hline
MAPG         & \multicolumn{1}{c|}{435.5595}                & \multicolumn{1}{c|}{77.6861}                & \multicolumn{1}{c|}{0.6195}                & \multicolumn{1}{c|}{1179}                & \multicolumn{1}{c|}{137.0363}                & \multicolumn{1}{c|}{78.4028}                & \multicolumn{1}{c|}{0.6875}                & 1130                \\ \hline
IPPO         & \multicolumn{1}{c|}{226.4010}                & \multicolumn{1}{c|}{58.2306}                & \multicolumn{1}{c|}{0.6641}                & \multicolumn{1}{c|}{1453}                & \multicolumn{1}{c|}{115.8968}                & \multicolumn{1}{c|}{52.4528}                & \multicolumn{1}{c|}{{\underline{ \textbf{0.4787}}}} & 1037                \\ \hline
PressLight   & \multicolumn{1}{c|}{{\underline{ \textbf{90.1724}}}}  & \multicolumn{1}{c|}{{\underline{ \textbf{13.2667}}}} & \multicolumn{1}{c|}{{\underline{ \textbf{0.4820}}}} & \multicolumn{1}{c|}{{\underline{ \textbf{1630}}}} & \multicolumn{1}{c|}{{\underline{ 91.8277}}}           & \multicolumn{1}{c|}{{\underline{ 13.7250}}}          & \multicolumn{1}{c|}{0.5153}                & {\underline{ 1608}}          \\ \hline
Network      & \multicolumn{8}{c|}{HangZhou4x4}                                                                                                                                                                                                                                                                                                                   \\ \hline
Simulator    & \multicolumn{4}{c|}{CityFlow}                                                                                                                                                      & \multicolumn{4}{c|}{SUMO}                                                                                                                                     \\ \hline
Metric       & \multicolumn{1}{c|}{Travel Time}             & \multicolumn{1}{c|}{Queue}                  & \multicolumn{1}{c|}{Delay}                 & \multicolumn{1}{c|}{Throughput}          & \multicolumn{1}{c|}{Travel Time}             & \multicolumn{1}{c|}{Queue}                  & \multicolumn{1}{c|}{Delay}                 & Throughput          \\ \hline
FixedTime(t\_fixed=10)      & \multicolumn{1}{c|}{689.0221}                & \multicolumn{1}{c|}{13.9837}                & \multicolumn{1}{c|}{0.9636}                & \multicolumn{1}{c|}{2385}                & \multicolumn{1}{c|}{535.0060}                & \multicolumn{1}{c|}{7.9405}                 & \multicolumn{1}{c|}{2.1130}                & 2495                \\ \hline
FixedTime(t\_fixed=30)      & \multicolumn{1}{c|}{\rebuttal{575.5565}}                & \multicolumn{1}{c|}{\rebuttal{12.3694}}                & \multicolumn{1}{c|}{\rebuttal{1.8439}}                & \multicolumn{1}{c|}{\rebuttal{2645}}                & \multicolumn{1}{c|}{\rebuttal{580.5826}}                & \multicolumn{1}{c|}{\rebuttal{10.3816}}                 & \multicolumn{1}{c|}{\rebuttal{2.4246}}                &  \rebuttal{2355}               \\ \hline
MaxPressurre & \multicolumn{1}{c|}{365.0634}                & \multicolumn{1}{c|}{3.5972}                 & \multicolumn{1}{c|}{1.7891}                & \multicolumn{1}{c|}{2928}                & \multicolumn{1}{c|}{350.4125}                & \multicolumn{1}{c|}{{\underline{ 1.2870}}}           & \multicolumn{1}{c|}{0.9678}                & {\underline{ 2732}} \\ \hline
SOTL         & \multicolumn{1}{c|}{354.1250}                & \multicolumn{1}{c|}{2.7229}                 & \multicolumn{1}{c|}{1.0208}                & \multicolumn{1}{c|}{2916}                & \multicolumn{1}{c|}{386.7881}                & \multicolumn{1}{c|}{3.3717}                 & \multicolumn{1}{c|}{1.4937}                & 2695                \\ \hline
IDQN         & \multicolumn{1}{c|}{{\underline{ \textbf{322.9068}}}} & \multicolumn{1}{c|}{{\underline{ \textbf{1.2141}}}}  & \multicolumn{1}{c|}{{\underline{ \textbf{0.0679}}}} & \multicolumn{1}{c|}{2929}                & \multicolumn{1}{c|}{{\underline{ \textbf{341.8509}}}} & \multicolumn{1}{c|}{{\underline{ \textbf{1.2097}}}}  & \multicolumn{1}{c|}{{\underline{ \textbf{0.0689}}}} & { 2730}          \\ \hline
MAPG         & \multicolumn{1}{c|}{782.3319}                & \multicolumn{1}{c|}{24.0474}                & \multicolumn{1}{c|}{0.1406}                & \multicolumn{1}{c|}{2331}                & \multicolumn{1}{c|}{436.2691}                & \multicolumn{1}{c|}{25.1415}                & \multicolumn{1}{c|}{0.2374}                & 1442                \\ \hline
IPPO         & \multicolumn{1}{c|}{727.9460}                & \multicolumn{1}{c|}{15.5786}                & \multicolumn{1}{c|}{0.1068}                & \multicolumn{1}{c|}{2306}                & \multicolumn{1}{c|}{418.0365}                & \multicolumn{1}{c|}{7.2328}                 & \multicolumn{1}{c|}{0.1560}                & 2496                \\ \hline
PressLight   & \multicolumn{1}{c|}{{\underline{ 329.7855}}}          & \multicolumn{1}{c|}{{\underline{ 1.6380}}}           & \multicolumn{1}{c|}{\underline{0.0841}}          & \multicolumn{1}{c|}{{\underline{ \textbf{2932}}}} & \multicolumn{1}{c|}{{354.0689}}          & \multicolumn{1}{c|}{1.4884}                 & \multicolumn{1}{c|}{{0.0803}}          & 2728                \\ \hline
Colight      & \multicolumn{1}{c|}{331.4348}                & \multicolumn{1}{c|}{1.7658}                 & \multicolumn{1}{c|}{0.0906}                & \multicolumn{1}{c|}{{\underline{ 2931}}}          & \multicolumn{1}{c|}{{\underline{343.4998}}}                        & \multicolumn{1}{c|}{1.3083}                       & \multicolumn{1}{c|}{{\underline{0.0714}}}                      &\underline{\textbf{2733}}                     \\ \hline
Network      & \multicolumn{8}{c|}{Manhattan7x28}                                                                                                                                                                                                                                                                                                                 \\ \hline
Simulator    & \multicolumn{4}{c|}{CityFlow}                                                                                                                                                      & \multicolumn{4}{c|}{SUMO}                                                                                                                                     \\ \hline
Metric       & \multicolumn{1}{c|}{Travel Time}             & \multicolumn{1}{c|}{Queue}                  & \multicolumn{1}{c|}{Delay}                 & \multicolumn{1}{c|}{Throughput}          & \multicolumn{1}{c|}{Travel Time}             & \multicolumn{1}{c|}{Queue}                  & \multicolumn{1}{c|}{Delay}                 & Throughput          \\ \hline
FixedTime(t\_fixed=10)      & \multicolumn{1}{c|}{1575.7847}               & \multicolumn{1}{c|}{20.7651}                & \multicolumn{1}{c|}{1.2703}                & \multicolumn{1}{c|}{7531}                & \multicolumn{1}{c|}{953.9797}                & \multicolumn{1}{c|}{19.3563}                & \multicolumn{1}{c|}{\underline{1.4353}}                & 2123                \\ \hline
FixedTime(t\_fixed=30)      & \multicolumn{1}{c|}{\rebuttal{1582.1030}}               & \multicolumn{1}{c|}{\rebuttal{22.3561}}                & \multicolumn{1}{c|}{\rebuttal{1.6659}}                & \multicolumn{1}{c|}{\rebuttal{7801}}                & \multicolumn{1}{c|}{\rebuttal{1144.1702}}                & \multicolumn{1}{c|}{\rebuttal{20.2811}}                & \multicolumn{1}{c|}{\rebuttal{1.8234}}                & \rebuttal{2379}                \\ \hline
MaxPressurre & \multicolumn{1}{c|}{\underline{1335.7877}}               & \multicolumn{1}{c|}{\underline{\textbf{17.3380}}}                 & \multicolumn{1}{c|}{1.3035}                & \multicolumn{1}{c|}{\underline{\textbf{9745}}}                & \multicolumn{1}{c|}{\underline{\textbf{797.4855}}}                & \multicolumn{1}{c|}{\underline{\textbf{15.2944}}}                & \multicolumn{1}{c|}{\underline{\textbf{1.3298}}}                & \underline{\textbf{4614}}                \\ \hline
SOTL         & \multicolumn{1}{c|}{1612.2468}               & \multicolumn{1}{c|}{23.6984}                & \multicolumn{1}{c|}{1.4968}                & \multicolumn{1}{c|}{8244}                & \multicolumn{1}{c|}{\underline{869.6206}}                & \multicolumn{1}{c|}{\underline{17.8344}}                & \multicolumn{1}{c|}{1.4543}                & \underline{3010}                \\ \hline
IDQN         & \multicolumn{1}{c|}{\underline{\textbf{{1319.4959}}}}               & \multicolumn{1}{c|}{\underline{17.4697}}                & \multicolumn{1}{c|}{\underline{\textbf{0.0916}}}                & \multicolumn{1}{c|}{9035}                & \multicolumn{1}{c|}{-}                        & \multicolumn{1}{c|}{-}                       & \multicolumn{1}{c|}{-}                      &                     -\\ \hline
MAPG         & \multicolumn{1}{c|}{1586.3388}                        & \multicolumn{1}{c|}{21.2705}                       & \multicolumn{1}{c|}{0.1159}                      & \multicolumn{1}{c|}{7481}                    & \multicolumn{1}{c|}{-}                        & \multicolumn{1}{c|}{-}                       & \multicolumn{1}{c|}{-}                      &    -                 \\ \hline
IPPO         & \multicolumn{1}{c|}{1468.4135}                        & \multicolumn{1}{c|}{19.4304}                       & \multicolumn{1}{c|}{0.1130}                      & \multicolumn{1}{c|}{8300}                    & \multicolumn{1}{c|}{-}                        & \multicolumn{1}{c|}{-}                       & \multicolumn{1}{c|}{-}                      & -                    \\ \hline
PressLight   & \multicolumn{1}{c|}{1338.7183}               & \multicolumn{1}{c|}{18.1332}                & \multicolumn{1}{c|}{\underline{0.0961}}                & \multicolumn{1}{c|}{\underline{9123}}                & \multicolumn{1}{c|}{-}                        & \multicolumn{1}{c|}{-}                       & \multicolumn{1}{c|}{-}                      &   -                  \\ \hline
CoLight      & \multicolumn{1}{c|}{1493.4200}                        & \multicolumn{1}{c|}{19.5024}                       & \multicolumn{1}{c|}{0.1007}                      & \multicolumn{1}{c|}{8287}                    & \multicolumn{1}{c|}{-}                        & \multicolumn{1}{c|}{-}                       & \multicolumn{1}{c|}{-}                      &   -                  \\ \hline
\end{tabular}
}
\footnotesize{\\ $^*$ Results shown as (-) indicate that no RL methods can be trained within acceptable time and resource \\in SUMO in Manhattan's road network.}
\end{table}



\subsection{Supplementary Results}
We conduct experiments on all nine datasets and also provide results of the best episode, full converge curves and standard deviations of the performance on the four datasets in full paper.

\subsubsection{Other comparison studies on datasets not shown in full paper}
Table~\ref{tab:sup-comparison-exp} shows the result of performance on the other five datasets. It shows \textbf{\rebuttal{\presslight}} and \textbf{\rebuttal{\dqn}} are the most stable algorithms at the most of the times. 

\subsubsection{Converge curve of Table~\ref{tab:comparison-experiment}}
\begin{figure*}[tbh]
\small
\centering
\resizebox{\columnwidth}{!}{
  \begin{tabular}{cc}
  \includegraphics[width=.48\linewidth]{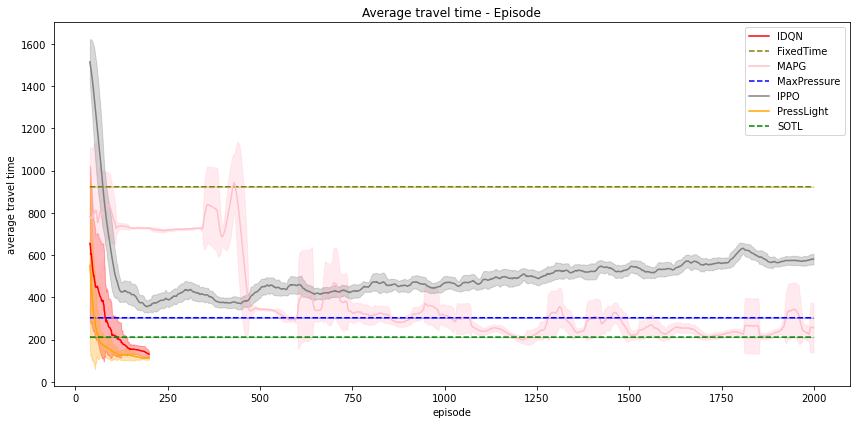} &
  \includegraphics[width=.48\linewidth]{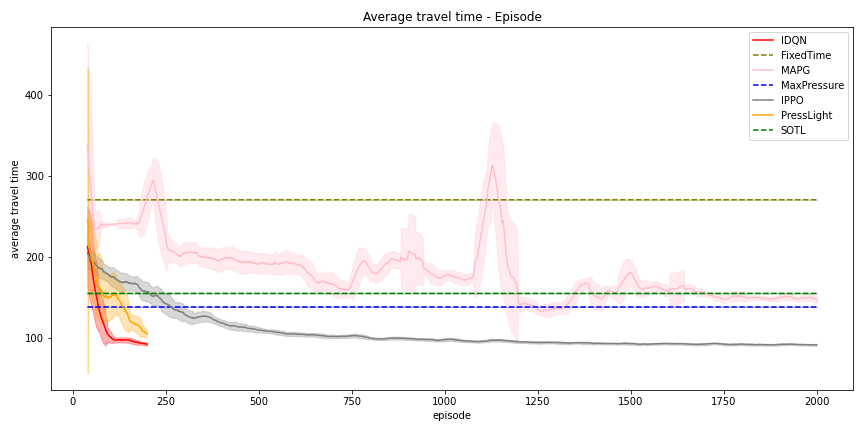} \\
  (a) \gridone in CityFlow & (b) \gridone in SUMO \\
    \includegraphics[width=.48\linewidth]{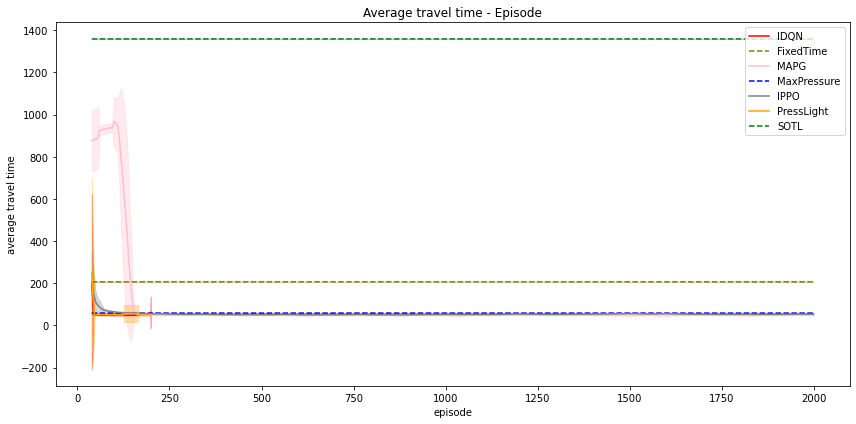} &
  \includegraphics[width=.48\linewidth]{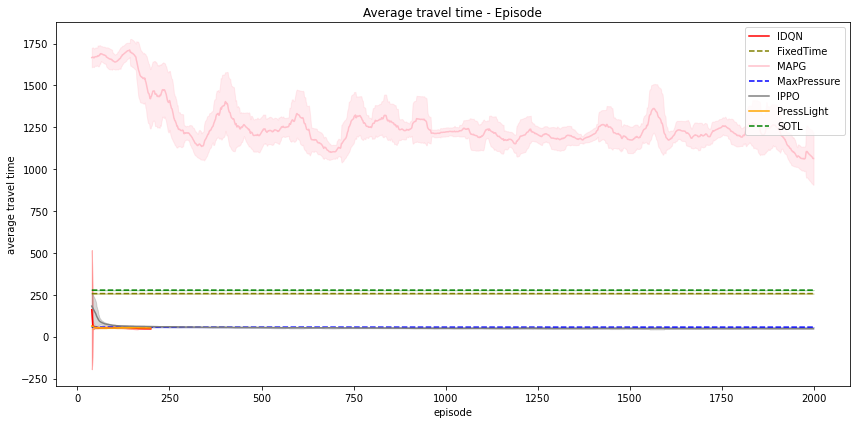} \\
  (c) \realone in CityFlow & (d) \realone in SUMO \\
    \includegraphics[width=.48\linewidth]{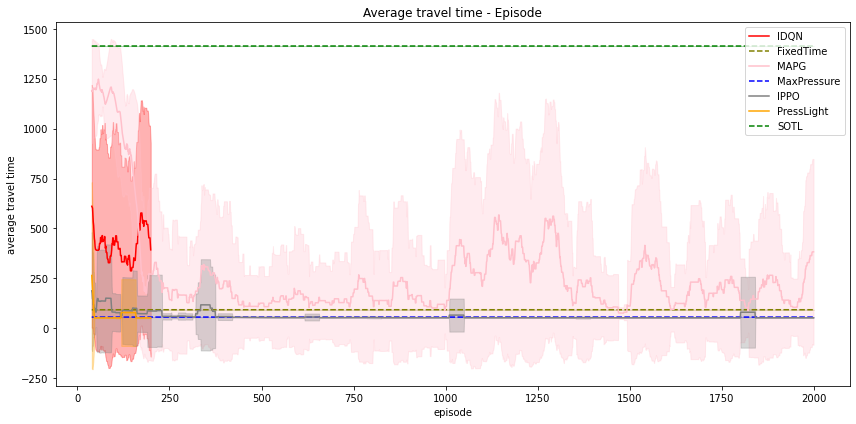} &
  \includegraphics[width=.48\linewidth]{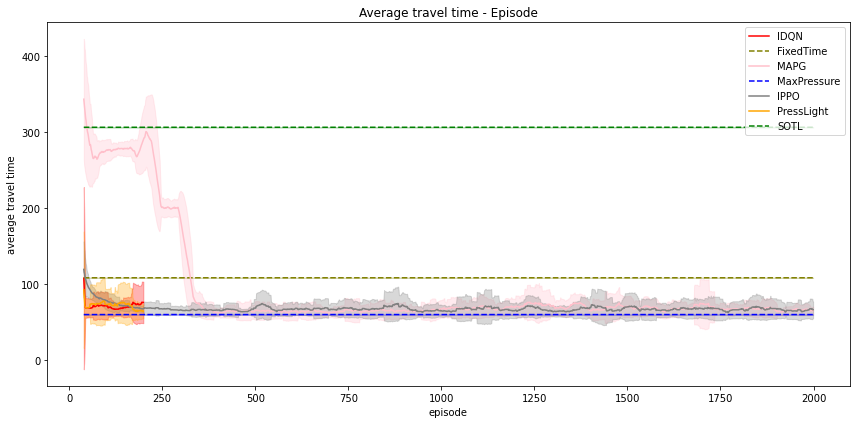} \\
  (e) \realthree in CityFlow & (f) \realthree in SUMO \\
    \includegraphics[width=.48\linewidth]{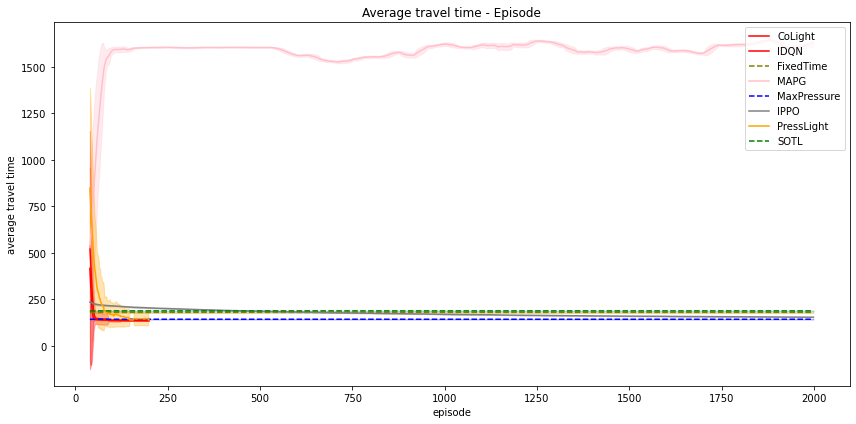} &
  \includegraphics[width=.48\linewidth]{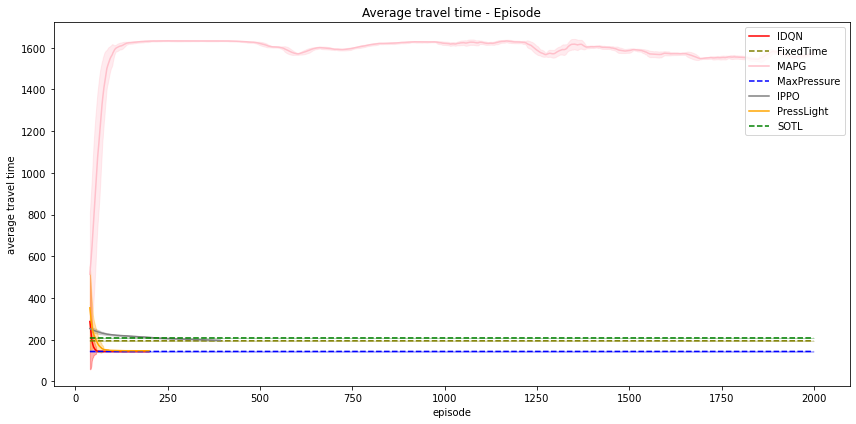} \\
  (g) \gridfour in CityFlow & (h) \gridfour in SUMO \\
  \end{tabular}}
 \caption{Full converge curve of Table~\ref{tab:comparison-experiment}}
    \label{fig:full AC algorithms}
\end{figure*}

Fig~\ref{fig:full AC algorithms} shows the full converge curve of 2000 episodes for \rebuttal{\ppo} and \rebuttal{\maddpg} agents. The result shows that comparing to Q-learning agents, Actor-Critic agents are hard to converge, on some large or complex datasets, converge time needed are more than ten times of Q-learning methods.

\subsubsection{Result of best episode}
Table~\ref{tab:best episode} gives the episode number of all datasets. It supports the conclusion that \textbf{\rebuttal{\presslight}}, followed by \textbf{\rebuttal{\dqn}}, have the best sample efficiency compared with other algorithms.

\begin{table}[tbh]
\centering
\caption{The episode of best results for different agents w.r.t. different methods}
\label{tab:best episode}
\small
\begin{tabular}{|c|c|c|c|c|c|}
\hline
Network                               & Simulator & IDQN & MAPG & IPPO & PressLight \\ \hline
\multirow{2}{*}{Cityflow1x1} & CityFlow  & 193  & 1205 & 185  & 197        \\ \cline{2-6} 
                                      & SUMO      & 187  & 188  & 194  & 185        \\ \hline
\multirow{2}{*}{Cologne1x1}           & CityFlow  & 95   & 1870 & 172  & 101        \\ \cline{2-6} 
                                      & SUMO      & 189  & 1992 & 346  & 193        \\ \hline
\multirow{2}{*}{Cologne1x3}           & CityFlow  & 133  & 597  & 1977 & 159        \\ \cline{2-6} 
                                      & SUMO      & 177  & 30   & 195  & 164        \\ \hline
\multirow{2}{*}{Grid4x4}              & CityFlow  & 163  & 6*   & 172  & 172        \\ \cline{2-6} 
                                      & SUMO      & 186  & 2*
                                    & 188  & 143       \\ \hline
\end{tabular}
\footnotesize{\\ $^*$ Though \rebuttal{\maddpg} and \rebuttal{\ppo} has the best results in the first few episodes, \\their performances are still worse than the other agents.}
\end{table}

\subsubsection{Performance on benchmark with standard deviations}
Table~\ref{tab:std} shows the standard deviation of the performance on the four datasets in full paper.

\begin{table}[tbh]
\centering
\caption{\rebuttal{The standard deviations of Table~\ref{tab:comparison-experiment}}}
\label{tab:std}
\resizebox{\textwidth}{!}{
\begin{tabular}{|c|cccccccc|}
\hline
Network &
  \multicolumn{8}{c|}{Cityflow 1x1} \\ \hline
Simulator &
  \multicolumn{4}{c|}{CityFlow} &
  \multicolumn{4}{c|}{SUMO} \\ \hline
Metric &
  \multicolumn{1}{c|}{Travel Time} &
  \multicolumn{1}{c|}{Queue} &
  \multicolumn{1}{c|}{Delay} &
  \multicolumn{1}{c|}{Throughput} &
  \multicolumn{1}{c|}{Travel Time} &
  \multicolumn{1}{c|}{Queue} &
  \multicolumn{1}{c|}{Delay} &
  Throughput \\ \hline
IDQN &
  \multicolumn{1}{c|}{9.474} &
  \multicolumn{1}{c|}{2.5995} &
  \multicolumn{1}{c|}{0.0221} &
  \multicolumn{1}{c|}{3.0496} &
  \multicolumn{1}{c|}{2.507} &
  \multicolumn{1}{c|}{0.6804} &
  \multicolumn{1}{c|}{0.4424} &
   0.0092\\ \hline
MAPG &
  \multicolumn{1}{c|}{10.746} &
  \multicolumn{1}{c|}{0.2149} &
  \multicolumn{1}{c|}{1.7188} &
  \multicolumn{1}{c|}{0.0152} &
  \multicolumn{1}{c|}{14.182} &
  \multicolumn{1}{c|}{0.6544} &
  \multicolumn{1}{c|}{0.4391} &
   0.0031\\ \hline
IPPO &
  \multicolumn{1}{c|}{23.020} &
  \multicolumn{1}{c|}{0.0011} &
  \multicolumn{1}{c|}{1.9196} &
  \multicolumn{1}{c|}{0.0055} &
  \multicolumn{1}{c|}{12.179} &
  \multicolumn{1}{c|}{0.0446} &
  \multicolumn{1}{c|}{7.7187} &
   0.015\\ \hline
PressLight &
  \multicolumn{1}{c|}{3.335} &
  \multicolumn{1}{c|}{1.334} &
  \multicolumn{1}{c|}{0.0127} &
  \multicolumn{1}{c|}{2.5884} &
  \multicolumn{1}{c|}{1.960} &
  \multicolumn{1}{c|}{3.261} &
  \multicolumn{1}{c|}{3.4475} &
   0.0141\\ \hline
FRAP &
  \multicolumn{1}{c|}{3.0118} &
  \multicolumn{1}{c|}{1.4411} &
  \multicolumn{1}{c|}{1.5028} &
  \multicolumn{1}{c|}{0.0106} &
  \multicolumn{1}{c|}{0.6933} &
  \multicolumn{1}{c|}{0.3848} &
  \multicolumn{1}{c|}{0.4273} &
  0.0095\\ \hline
Network &
  \multicolumn{8}{c|}{Cologne 1x1} \\ \hline
Simulator &
  \multicolumn{4}{c|}{CityFlow} &
  \multicolumn{4}{c|}{SUMO} \\ \hline
Metric &
  \multicolumn{1}{c|}{Travel Time} &
  \multicolumn{1}{c|}{Queue} &
  \multicolumn{1}{c|}{Delay} &
  \multicolumn{1}{c|}{Throughput} &
  \multicolumn{1}{c|}{Travel Time} &
  \multicolumn{1}{c|}{Queue} &
  \multicolumn{1}{c|}{Delay} &
  Throughput \\ \hline
IDQN &
  \multicolumn{1}{c|}{2.100} &
  \multicolumn{1}{c|}{0.6913} &
  \multicolumn{1}{c|}{0.7502} &
  \multicolumn{1}{c|}{0.0095} &
  \multicolumn{1}{c|}{0.144} &
  \multicolumn{1}{c|}{0.6317} &
  \multicolumn{1}{c|}{0.3986} &
   0.0087\\ \hline
MAPG &
  \multicolumn{1}{c|}{2.186} &
  \multicolumn{1}{c|}{0.0359} &
  \multicolumn{1}{c|}{0.2659} &
  \multicolumn{1}{c|}{0.0072} &
  \multicolumn{1}{c|}{12.283} &
  \multicolumn{1}{c|}{0.0088} &
  \multicolumn{1}{c|}{0.0744} &
   0.0001\\ \hline
IPPO &
  \multicolumn{1}{c|}{28.048} &
  \multicolumn{1}{c|}{0.0003} &
  \multicolumn{1}{c|}{0.0495} &
  \multicolumn{1}{c|}{0.0015} &
  \multicolumn{1}{c|}{0.950} &
  \multicolumn{1}{c|}{0.0013} &
  \multicolumn{1}{c|}{0.1982} &
   0.0008\\ \hline
PressLight &
  \multicolumn{1}{c|}{2.274} &
  \multicolumn{1}{c|}{0.3753} &
  \multicolumn{1}{c|}{0.3161} &
  \multicolumn{1}{c|}{0.0066} &
  \multicolumn{1}{c|}{0.767} &
  \multicolumn{1}{c|}{0.6011} &
  \multicolumn{1}{c|}{0.6375} &
   0.0106\\ \hline
Network &
  \multicolumn{8}{c|}{Cologne 1x3} \\ \hline
Simulator &
  \multicolumn{4}{c|}{CityFlow} &
  \multicolumn{4}{c|}{SUMO} \\ \hline
Metric &
  \multicolumn{1}{c|}{Travel Time} &
  \multicolumn{1}{c|}{Queue} &
  \multicolumn{1}{c|}{Delay} &
  \multicolumn{1}{c|}{Throughput} &
  \multicolumn{1}{c|}{Travel Time} &
  \multicolumn{1}{c|}{Queue} &
  \multicolumn{1}{c|}{Delay} &
  Throughput \\ \hline
IDQN &
  \multicolumn{1}{c|}{5.891} &
  \multicolumn{1}{c|}{0.9075} &
  \multicolumn{1}{c|}{0.1778} &
  \multicolumn{1}{c|}{0.0053} &
  \multicolumn{1}{c|}{0.875} &
  \multicolumn{1}{c|}{0.0711} &
  \multicolumn{1}{c|}{0.0111} &
   0.0017\\ \hline
MAPG &
  \multicolumn{1}{c|}{2.883} &
  \multicolumn{1}{c|}{0.0637} &
  \multicolumn{1}{c|}{0.1467} &
  \multicolumn{1}{c|}{0.0047} &
  \multicolumn{1}{c|}{15.494} &
  \multicolumn{1}{c|}{3.5459} &
  \multicolumn{1}{c|}{0.5764} &
   0.0134\\ \hline
IPPO &
  \multicolumn{1}{c|}{0.950} &
  \multicolumn{1}{c|}{0.0} &
  \multicolumn{1}{c|}{0.0159} &
  \multicolumn{1}{c|}{0.002} &
  \multicolumn{1}{c|}{4.107} &
  \multicolumn{1}{c|}{0.0002} &
  \multicolumn{1}{c|}{0.1743} &
   0.0078\\ \hline
PressLight &
  \multicolumn{1}{c|}{0.619} &
  \multicolumn{1}{c|}{0.124} &
  \multicolumn{1}{c|}{0.0405} &
  \multicolumn{1}{c|}{0.0026} &
  \multicolumn{1}{c|}{4.982} &
  \multicolumn{1}{c|}{2.3948} &
  \multicolumn{1}{c|}{0.798} &
   0.0123\\ \hline
Network &
  \multicolumn{8}{c|}{Grid4x4} \\ \hline
Simulator &
  \multicolumn{4}{c|}{CityFlow} &
  \multicolumn{4}{c|}{SUMO} \\ \hline
Metric &
  \multicolumn{1}{c|}{Travel Time} &
  \multicolumn{1}{c|}{Queue} &
  \multicolumn{1}{c|}{Delay} &
  \multicolumn{1}{c|}{Throughput} &
  \multicolumn{1}{c|}{Travel Time} &
  \multicolumn{1}{c|}{Queue} &
  \multicolumn{1}{c|}{Delay} &
  Throughput \\ \hline
IDQN &
  \multicolumn{1}{c|}{2.006} &
  \multicolumn{1}{c|}{0.0117} &
  \multicolumn{1}{c|}{0.0006} &
  \multicolumn{1}{c|}{0.0} &
  \multicolumn{1}{c|}{0.795} &
  \multicolumn{1}{c|}{0.1733} &
  \multicolumn{1}{c|}{0.0098} &
   0.0005\\ \hline
MAPG &
  \multicolumn{1}{c|}{13.330} &
  \multicolumn{1}{c|}{1.1346} &
  \multicolumn{1}{c|}{0.071} &
  \multicolumn{1}{c|}{0.0044} &
  \multicolumn{1}{c|}{5.420} &
  \multicolumn{1}{c|}{3.6906} &
  \multicolumn{1}{c|}{0.2473} &
   0.0049\\ \hline
IPPO &
  \multicolumn{1}{c|}{2.432} &
  \multicolumn{1}{c|}{0.0003} &
  \multicolumn{1}{c|}{0.0495} &
  \multicolumn{1}{c|}{0.0015} &
  \multicolumn{1}{c|}{3.465} &
  \multicolumn{1}{c|}{0.0004} &
  \multicolumn{1}{c|}{0.0346} &
   0.0004\\ \hline
PressLight &
  \multicolumn{1}{c|}{0.880} &
  \multicolumn{1}{c|}{0.0491} &
  \multicolumn{1}{c|}{0.0022} &
  \multicolumn{1}{c|}{0.0} &
  \multicolumn{1}{c|}{0.636} &
  \multicolumn{1}{c|}{0.1045} &
  \multicolumn{1}{c|}{0.0132} &
   0.0003\\ \hline
CoLight &
  \multicolumn{1}{c|}{0.8315} &
  \multicolumn{1}{c|}{0.0161} &
  \multicolumn{1}{c|}{0.0009} &
  \multicolumn{1}{c|}{0.0} &
  \multicolumn{1}{c|}{1.1633} &
  \multicolumn{1}{c|}{0.621} &
  \multicolumn{1}{c|}{0.0384} &
   0.0017\\ \hline
\end{tabular}
}
\end{table}



\subsection{Extension to Other Simulators}
\ours is a cross-simulator library for traffic control tasks. Currently, we support the most commonly used CityFlow and SUMO simulators, and our library is open to other new simulation environments. CBEngine is a new simulator that served as the simulation environment in the KDD Cup 2021 City Brain Challenge~\footnote{\url{http://www.yunqiacademy.org/poster}} and is designed for executing traffic control tasks on large traffic networks. We integrate this new simulator into our traffic control framework to extend \ours's usage in other simulation environments. We show the result of \rebuttal{\maxp}, \rebuttal{\sotl}, \rebuttal{\ft}, and \rebuttal{\dqn}' performance under CBEngine in Table~\ref{tab:cbengine}

\begin{table}[t!]
\centering
\caption{Performance on CBEngine simulator}
\label{tab:cbengine}
\begin{tabular}{|c|c|c|c|c|}
\hline
CityFlow1x1      & FixedTime & MaxPressure & SOTL    & IDQN    \\ \hline
Avg. Travel Time & 654.4848  & 150.7677    & 96.0025 & 84.5404 \\ \hline
\end{tabular}
\end{table}

\subsection{Hyperparameters}
Table~\ref{tab:hyp-params} provides the parameters of each algorithm, training environment, and hardware parameters on the server.



\begin{table}[tbh]
\centering
\caption{Hyperparameters of models, servers and training}
\label{tab:hyp-params}
\resizebox{\textwidth}{!}{
\begin{tabular}{|ccccccccc|}
\hline
\multicolumn{9}{|c|}{\textbf{Model Parameters}}                                                                                                                                                                                                                                                                                                                                 \\ \hline
\multicolumn{1}{|c|}{\multirow{3}{*}{\textbf{FixedTime}}}      & \multicolumn{1}{c|}{t\_fixed}            & \multicolumn{1}{c|}{30}            & \multicolumn{1}{c|}{buffer\_size}         & \multicolumn{1}{c|}{0}             & \multicolumn{1}{c|}{learning\_rate}       & \multicolumn{1}{c|}{0}         & \multicolumn{1}{c|}{learning\_start}               & 0             \\ \cline{2-9} 
\multicolumn{1}{|c|}{}                                       & \multicolumn{1}{c|}{update\_model\_rate} & \multicolumn{1}{c|}{0}             & \multicolumn{1}{c|}{update\_target\_rate} & \multicolumn{1}{c|}{0}             & \multicolumn{1}{c|}{save\_rate}           & \multicolumn{1}{c|}{0}         & \multicolumn{1}{c|}{train\_model}                  & false         \\ \cline{2-9} 
\multicolumn{1}{|c|}{}                                       & \multicolumn{1}{c|}{test\_model}         & \multicolumn{1}{c|}{true}          & \multicolumn{1}{c|}{one\_hot}             & \multicolumn{1}{c|}{false}         & \multicolumn{1}{c|}{phase}                & \multicolumn{1}{c|}{false}     & \multicolumn{1}{c|}{episodes}                      & 1             \\ \hline
\multicolumn{1}{|c|}{\multirow{3}{*}{\textbf{MaxPressure}}} & \multicolumn{1}{c|}{t\_min}              & \multicolumn{1}{c|}{10}            & \multicolumn{1}{c|}{buffer\_size}         & \multicolumn{1}{c|}{0}             & \multicolumn{1}{c|}{learning\_rate}       & \multicolumn{1}{c|}{0}         & \multicolumn{1}{c|}{learning\_start}               & 0             \\ \cline{2-9} 
\multicolumn{1}{|c|}{}                                       & \multicolumn{1}{c|}{update\_model\_rate} & \multicolumn{1}{c|}{0}             & \multicolumn{1}{c|}{update\_target\_rate} & \multicolumn{1}{c|}{0}             & \multicolumn{1}{c|}{save\_rate}           & \multicolumn{1}{c|}{0}         & \multicolumn{1}{c|}{train\_model}                  & false         \\ \cline{2-9} 
\multicolumn{1}{|c|}{}                                       & \multicolumn{1}{c|}{test\_model}         & \multicolumn{1}{c|}{true}          & \multicolumn{1}{c|}{one\_hot}             & \multicolumn{1}{c|}{false}         & \multicolumn{1}{c|}{phase}                & \multicolumn{1}{c|}{false}     & \multicolumn{1}{c|}{episodes}                      & 1             \\ \hline
\multicolumn{1}{|c|}{\multirow{4}{*}{\textbf{SOTL}}}         & \multicolumn{1}{c|}{t\_min}              & \multicolumn{1}{c|}{5}             & \multicolumn{1}{c|}{min\_green\_vehicle}  & \multicolumn{1}{c|}{3}             & \multicolumn{1}{c|}{max\_red\_vehicle}    & \multicolumn{1}{c|}{6}         & \multicolumn{1}{c|}{buffer\_size}                  & 0             \\ \cline{2-9} 
\multicolumn{1}{|c|}{}                                       & \multicolumn{1}{c|}{learning\_rate}      & \multicolumn{1}{c|}{0}             & \multicolumn{1}{c|}{learning\_start}      & \multicolumn{1}{c|}{0}             & \multicolumn{1}{c|}{update\_model\_rate}  & \multicolumn{1}{c|}{0}         & \multicolumn{1}{c|}{update\_target\_rate}          & 0             \\ \cline{2-9} 
\multicolumn{1}{|c|}{}                                       & \multicolumn{1}{c|}{save\_rate}          & \multicolumn{1}{c|}{0}             & \multicolumn{1}{c|}{train\_model}         & \multicolumn{1}{c|}{false}         & \multicolumn{1}{c|}{test\_model}          & \multicolumn{1}{c|}{true}      & \multicolumn{1}{c|}{one\_hot}                      & false         \\ \cline{2-9} 
\multicolumn{1}{|c|}{}                                       & \multicolumn{1}{c|}{phase}               & \multicolumn{1}{c|}{false}         & \multicolumn{1}{c|}{episodes}             & \multicolumn{1}{c|}{1}             & \multicolumn{1}{c|}{}                     & \multicolumn{1}{c|}{}          & \multicolumn{1}{c|}{}                              &               \\ \hline
\multicolumn{1}{|c|}{\multirow{6}{*}{\textbf{IDQN}}}         & \multicolumn{1}{c|}{learning\_rate}      & \multicolumn{1}{c|}{0.001}         & \multicolumn{1}{c|}{learning\_start}      & \multicolumn{1}{c|}{1000}          & \multicolumn{1}{c|}{graphic}              & \multicolumn{1}{c|}{false}     & \multicolumn{1}{c|}{buffer\_size}                  & 5000          \\ \cline{2-9} 
\multicolumn{1}{|c|}{}                                       & \multicolumn{1}{c|}{batch\_size}         & \multicolumn{1}{c|}{64}            & \multicolumn{1}{c|}{episodes}             & \multicolumn{1}{c|}{200}           & \multicolumn{1}{c|}{epsilon}              & \multicolumn{1}{c|}{0.1}       & \multicolumn{1}{c|}{epsilon\_decay}                & 0.995         \\ \cline{2-9} 
\multicolumn{1}{|c|}{}                                       & \multicolumn{1}{c|}{epsilon\_min}        & \multicolumn{1}{c|}{0.01}          & \multicolumn{1}{c|}{update\_model\_rate}  & \multicolumn{1}{c|}{1}             & \multicolumn{1}{c|}{update\_target\_rate} & \multicolumn{1}{c|}{10}        & \multicolumn{1}{c|}{save\_rate}                    & 20            \\ \cline{2-9} 
\multicolumn{1}{|c|}{}                                       & \multicolumn{1}{c|}{one\_hot}            & \multicolumn{1}{c|}{true}          & \multicolumn{1}{c|}{phase}                & \multicolumn{1}{c|}{true}          & \multicolumn{1}{c|}{gamma}                & \multicolumn{1}{c|}{0.95}      & \multicolumn{1}{c|}{steps}                         & 3600          \\ \cline{2-9} 
\multicolumn{1}{|c|}{}                                       & \multicolumn{1}{c|}{test\_steps}         & \multicolumn{1}{c|}{3600}          & \multicolumn{1}{c|}{vehicle\_max}         & \multicolumn{1}{c|}{1}             & \multicolumn{1}{c|}{grad\_clip}           & \multicolumn{1}{c|}{5}         & \multicolumn{1}{c|}{train\_model}                  & false         \\ \cline{2-9} 
\multicolumn{1}{|c|}{}                                       & \multicolumn{1}{c|}{test\_model}         & \multicolumn{1}{c|}{true}          & \multicolumn{1}{c|}{action\_interval}     & \multicolumn{1}{c|}{10}            & \multicolumn{1}{c|}{}                     & \multicolumn{1}{c|}{}          & \multicolumn{1}{c|}{}                              &               \\ \hline
\multicolumn{1}{|c|}{\multirow{6}{*}{\textbf{MAPG}}}         & \multicolumn{1}{c|}{tau}                 & \multicolumn{1}{c|}{0.01}           & \multicolumn{1}{c|}{learning\_rate}       & \multicolumn{1}{c|}{0.001}         & \multicolumn{1}{c|}{learning\_start}      & \multicolumn{1}{c|}{5000}      & \multicolumn{1}{c|}{graphic}                       & false         \\ \cline{2-9} 
\multicolumn{1}{|c|}{}                                       & \multicolumn{1}{c|}{buffer\_size}        & \multicolumn{1}{c|}{10000}         & \multicolumn{1}{c|}{batch\_size}          & \multicolumn{1}{c|}{256}           & \multicolumn{1}{c|}{episodes}             & \multicolumn{1}{c|}{2000}      & \multicolumn{1}{c|}{epsilon}                       & 0.5           \\ \cline{2-9} 
\multicolumn{1}{|c|}{}                                       & \multicolumn{1}{c|}{epsilon\_decay}      & \multicolumn{1}{c|}{0.99}          & \multicolumn{1}{c|}{epsilon\_min}         & \multicolumn{1}{c|}{0.01}          & \multicolumn{1}{c|}{update\_model\_rate}  & \multicolumn{1}{c|}{30}        & \multicolumn{1}{c|}{update\_target\_rate}          & 30            \\ \cline{2-9} 
\multicolumn{1}{|c|}{}                                       & \multicolumn{1}{c|}{save\_rate}          & \multicolumn{1}{c|}{1000}          & \multicolumn{1}{c|}{one\_hot}             & \multicolumn{1}{c|}{FLASE}         & \multicolumn{1}{c|}{phase}                & \multicolumn{1}{c|}{false}     & \multicolumn{1}{c|}{gamma}                         & 0.95          \\ \cline{2-9} 
\multicolumn{1}{|c|}{}                                       & \multicolumn{1}{c|}{steps}               & \multicolumn{1}{c|}{3600}          & \multicolumn{1}{c|}{test\_steps}          & \multicolumn{1}{c|}{3600}          & \multicolumn{1}{c|}{vehicle\_max}         & \multicolumn{1}{c|}{1}         & \multicolumn{1}{c|}{grad\_clip}                    & 5             \\ \cline{2-9} 
\multicolumn{1}{|c|}{}                                       & \multicolumn{1}{c|}{train\_model}        & \multicolumn{1}{c|}{false}         & \multicolumn{1}{c|}{test\_model}          & \multicolumn{1}{c|}{true}          & \multicolumn{1}{c|}{action\_interval}     & \multicolumn{1}{c|}{10}        & \multicolumn{1}{c|}{}                              &               \\ \hline
\multicolumn{1}{|c|}{\multirow{6}{*}{\textbf{IPPO}}}         & \multicolumn{1}{c|}{learning\_rate}      & \multicolumn{1}{c|}{0.0001}        & \multicolumn{1}{c|}{learning\_start}      & \multicolumn{1}{c|}{-1}            & \multicolumn{1}{c|}{graphic}              & \multicolumn{1}{c|}{false}     & \multicolumn{1}{c|}{buffer\_size}                  & 5000          \\ \cline{2-9} 
\multicolumn{1}{|c|}{}                                       & \multicolumn{1}{c|}{batch\_size}         & \multicolumn{1}{c|}{64}            & \multicolumn{1}{c|}{episodes}             & \multicolumn{1}{c|}{2000}          & \multicolumn{1}{c|}{epsilon}              & \multicolumn{1}{c|}{0.1}       & \multicolumn{1}{c|}{epsilon\_decay}                & 0.995         \\ \cline{2-9} 
\multicolumn{1}{|c|}{}                                       & \multicolumn{1}{c|}{epsilon\_min}        & \multicolumn{1}{c|}{0.01}          & \multicolumn{1}{c|}{update\_model\_rate}  & \multicolumn{1}{c|}{1}             & \multicolumn{1}{c|}{update\_target\_rate} & \multicolumn{1}{c|}{10}        & \multicolumn{1}{c|}{save\_rate}                    & 1000          \\ \cline{2-9} 
\multicolumn{1}{|c|}{}                                       & \multicolumn{1}{c|}{one\_hot}            & \multicolumn{1}{c|}{true}          & \multicolumn{1}{c|}{phase}                & \multicolumn{1}{c|}{true}          & \multicolumn{1}{c|}{gamma}                & \multicolumn{1}{c|}{0.95}      & \multicolumn{1}{c|}{steps}                         & 3600          \\ \cline{2-9} 
\multicolumn{1}{|c|}{}                                       & \multicolumn{1}{c|}{test\_steps}         & \multicolumn{1}{c|}{3600}          & \multicolumn{1}{c|}{vehicle\_max}         & \multicolumn{1}{c|}{1}             & \multicolumn{1}{c|}{grad\_clip}           & \multicolumn{1}{c|}{5}         & \multicolumn{1}{c|}{train\_model}                  & false         \\ \cline{2-9} 
\multicolumn{1}{|c|}{}                                       & \multicolumn{1}{c|}{test\_model}         & \multicolumn{1}{c|}{true}          & \multicolumn{1}{c|}{action\_interval}     & \multicolumn{1}{c|}{10}            & \multicolumn{1}{c|}{}                     & \multicolumn{1}{c|}{}          & \multicolumn{1}{c|}{}                              &               \\ \hline
\multicolumn{1}{|c|}{\multirow{7}{*}{\textbf{PressLight}}}   & \multicolumn{1}{c|}{d\_dense}            & \multicolumn{1}{c|}{20}            & \multicolumn{1}{c|}{n\_layer}             & \multicolumn{1}{c|}{2}             & \multicolumn{1}{c|}{normal\_factor}       & \multicolumn{1}{c|}{20}        & \multicolumn{1}{c|}{patience}                      & 10            \\ \cline{2-9} 
\multicolumn{1}{|c|}{}                                       & \multicolumn{1}{c|}{learning\_rate}      & \multicolumn{1}{c|}{0.001}         & \multicolumn{1}{c|}{learning\_start}      & \multicolumn{1}{c|}{1000}          & \multicolumn{1}{c|}{graphic}              & \multicolumn{1}{c|}{false}     & \multicolumn{1}{c|}{buffer\_size}                  & 5000          \\ \cline{2-9} 
\multicolumn{1}{|c|}{}                                       & \multicolumn{1}{c|}{batch\_size}         & \multicolumn{1}{c|}{64}            & \multicolumn{1}{c|}{episodes}             & \multicolumn{1}{c|}{200}           & \multicolumn{1}{c|}{epsilon}              & \multicolumn{1}{c|}{0.1}       & \multicolumn{1}{c|}{epsilon\_decay}                & 0.995         \\ \cline{2-9} 
\multicolumn{1}{|c|}{}                                       & \multicolumn{1}{c|}{epsilon\_min}        & \multicolumn{1}{c|}{0.01}          & \multicolumn{1}{c|}{update\_model\_rate}  & \multicolumn{1}{c|}{1}             & \multicolumn{1}{c|}{update\_target\_rate} & \multicolumn{1}{c|}{10}        & \multicolumn{1}{c|}{save\_rate}                    & 20            \\ \cline{2-9} 
\multicolumn{1}{|c|}{}                                       & \multicolumn{1}{c|}{one\_hot}            & \multicolumn{1}{c|}{true}          & \multicolumn{1}{c|}{phase}                & \multicolumn{1}{c|}{true}          & \multicolumn{1}{c|}{gamma}                & \multicolumn{1}{c|}{0.95}      & \multicolumn{1}{c|}{steps}                         & 3600          \\ \cline{2-9} 
\multicolumn{1}{|c|}{}                                       & \multicolumn{1}{c|}{test\_steps}         & \multicolumn{1}{c|}{3600}          & \multicolumn{1}{c|}{vehicle\_max}         & \multicolumn{1}{c|}{1}             & \multicolumn{1}{c|}{grad\_clip}           & \multicolumn{1}{c|}{5}         & \multicolumn{1}{c|}{train\_model}                  & false         \\ \cline{2-9} 
\multicolumn{1}{|c|}{}                                       & \multicolumn{1}{c|}{test\_model}         & \multicolumn{1}{c|}{true}          & \multicolumn{1}{c|}{action\_interval}     & \multicolumn{1}{c|}{10}            & \multicolumn{1}{c|}{}                     & \multicolumn{1}{c|}{}          & \multicolumn{1}{c|}{}                              &               \\ \hline
\multicolumn{1}{|c|}{\multirow{6}{*}{\textbf{FRAP}}}         & \multicolumn{1}{c|}{d\_dense}            & \multicolumn{1}{c|}{20}            & \multicolumn{1}{c|}{n\_layer}             & \multicolumn{1}{c|}{2}             & \multicolumn{1}{c|}{one\_hot}             & \multicolumn{1}{c|}{false}     & \multicolumn{1}{c|}{phase}                         & true                  \\ \cline{2-9} 
\multicolumn{1}{|c|}{}                                       & \multicolumn{1}{c|}{learning\_rate}      & \multicolumn{1}{c|}{0.001}         & \multicolumn{1}{c|}{learning\_start}      & \multicolumn{1}{c|}{1000}          & \multicolumn{1}{c|}{graphic}              & \multicolumn{1}{c|}{false}     & \multicolumn{1}{c|}{buffer\_size}                  & 5000                  \\ \cline{2-9} 
\multicolumn{1}{|c|}{}                                       & \multicolumn{1}{c|}{rotation}            & \multicolumn{1}{c|}{true}          & \multicolumn{1}{c|}{conflict\_matrix}     & \multicolumn{1}{c|}{true}          & \multicolumn{1}{c|}{merge}                & \multicolumn{1}{c|}{multiply}  & \multicolumn{1}{c|}{demand\_shape}                 & 1                     \\ \cline{2-9} 
\multicolumn{1}{|c|}{}                                       & \multicolumn{1}{c|}{batch\_size}         & \multicolumn{1}{c|}{64}            & \multicolumn{1}{c|}{episodes}             & \multicolumn{1}{c|}{200}           & \multicolumn{1}{c|}{epsilon}              & \multicolumn{1}{c|}{0.1}       & \multicolumn{1}{c|}{epsilon\_decay}                & 0.995                 \\ \cline{2-9} 
\multicolumn{1}{|c|}{}                                       & \multicolumn{1}{c|}{epsilon\_min}        & \multicolumn{1}{c|}{0.01}          & \multicolumn{1}{c|}{update\_model\_rate}  & \multicolumn{1}{c|}{1}             & \multicolumn{1}{c|}{update\_target\_rate} & \multicolumn{1}{c|}{10}        & \multicolumn{1}{c|}{save\_rate}                    & 20                    \\ \cline{2-9} 
\multicolumn{1}{|c|}{}                                       & \multicolumn{1}{c|}{test\_steps}         & \multicolumn{1}{c|}{3600}          & \multicolumn{1}{c|}{test\_model}          & \multicolumn{1}{c|}{true}          & \multicolumn{1}{c|}{action\_interval}     & \multicolumn{1}{c|}{10}        & \multicolumn{1}{c|}{train\_model}                  & true                  \\ \hline
\multicolumn{1}{|c|}{\multirow{6}{*}{\textbf{MPLight}}}      & \multicolumn{1}{c|}{d\_dense}            & \multicolumn{1}{c|}{20}            & \multicolumn{1}{c|}{n\_layer}             & \multicolumn{1}{c|}{2}             & \multicolumn{1}{c|}{one\_hot}             & \multicolumn{1}{c|}{false}     & \multicolumn{1}{c|}{phase}                         & true                  \\ \cline{2-9} 
\multicolumn{1}{|c|}{}                                       & \multicolumn{1}{c|}{learning\_rate}      & \multicolumn{1}{c|}{0.001}         & \multicolumn{1}{c|}{learning\_start}      & \multicolumn{1}{c|}{-1}            & \multicolumn{1}{c|}{graphic}              & \multicolumn{1}{c|}{false}     & \multicolumn{1}{c|}{buffer\_size}                  & 10000                 \\ \cline{2-9} 
\multicolumn{1}{|c|}{}                                       & \multicolumn{1}{c|}{rotation}            & \multicolumn{1}{c|}{true}          & \multicolumn{1}{c|}{conflict\_matrix}     & \multicolumn{1}{c|}{true}          & \multicolumn{1}{c|}{merge}                & \multicolumn{1}{c|}{multiply}  & \multicolumn{1}{c|}{demand\_shape}                 & 1                     \\ \cline{2-9} 
\multicolumn{1}{|c|}{}                                       & \multicolumn{1}{c|}{batch\_size}         & \multicolumn{1}{c|}{32}            & \multicolumn{1}{c|}{episodes}             & \multicolumn{1}{c|}{200}           & \multicolumn{1}{c|}{eps\_start}           & \multicolumn{1}{c|}{1}         & \multicolumn{1}{c|}{eps\_end}                      & 0                     \\ \cline{2-9} 
\multicolumn{1}{|c|}{}                                       & \multicolumn{1}{c|}{eps\_decay}          & \multicolumn{1}{c|}{220}           & \multicolumn{1}{c|}{target\_update}       & \multicolumn{1}{c|}{500}           & \multicolumn{1}{c|}{gamma}                & \multicolumn{1}{c|}{0.99}      & \multicolumn{1}{c|}{save\_rate}                    & 20                    \\ \cline{2-9} 
\multicolumn{1}{|c|}{}                                       & \multicolumn{1}{c|}{test\_steps}         & \multicolumn{1}{c|}{3600}          & \multicolumn{1}{c|}{test\_model}          & \multicolumn{1}{c|}{true}          & \multicolumn{1}{c|}{action\_interval}     & \multicolumn{1}{c|}{10}        & \multicolumn{1}{c|}{train\_model}                  & true                  \\ \hline
\multicolumn{1}{|c|}{\multirow{8}{*}{\textbf{CoLight}}}      & \multicolumn{1}{c|}{neighbor\_num}       & \multicolumn{1}{c|}{4}             & \multicolumn{1}{c|}{neighbor\_edge\_num}  & \multicolumn{1}{c|}{4}             & \multicolumn{1}{c|}{n\_layer}             & \multicolumn{1}{c|}{1}         & \multicolumn{1}{c|}{input\_dim}                    & {[}128,128{]} \\ \cline{2-9} 
\multicolumn{1}{|c|}{}                                       & \multicolumn{1}{c|}{output\_dim}         & \multicolumn{1}{c|}{{[}128,128{]}} & \multicolumn{1}{c|}{node\_emb\_dim}       & \multicolumn{1}{c|}{{[}128,128{]}} & \multicolumn{1}{c|}{num\_heads}           & \multicolumn{1}{c|}{{[}5,5{]}} & \multicolumn{1}{c|}{node\_layer\_dims\_each\_head} & {[}16,16{]}   \\ \cline{2-9} 
\multicolumn{1}{|c|}{}                                       & \multicolumn{1}{c|}{output\_layers}      & \multicolumn{1}{c|}{{[}{]}}        & \multicolumn{1}{c|}{learning\_rate}       & \multicolumn{1}{c|}{0.001}         & \multicolumn{1}{c|}{learning\_start}      & \multicolumn{1}{c|}{1000}      & \multicolumn{1}{c|}{graphic}                       & true          \\ \cline{2-9} 
\multicolumn{1}{|c|}{}                                       & \multicolumn{1}{c|}{buffer\_size}        & \multicolumn{1}{c|}{5000}          & \multicolumn{1}{c|}{batch\_size}          & \multicolumn{1}{c|}{64}            & \multicolumn{1}{c|}{episodes}             & \multicolumn{1}{c|}{200}       & \multicolumn{1}{c|}{epsilon}                       & 0.8           \\ \cline{2-9} 
\multicolumn{1}{|c|}{}                                       & \multicolumn{1}{c|}{epsilon\_decay}      & \multicolumn{1}{c|}{0.9995}        & \multicolumn{1}{c|}{epsilon\_min}         & \multicolumn{1}{c|}{0.01}          & \multicolumn{1}{c|}{update\_model\_rate}  & \multicolumn{1}{c|}{1}         & \multicolumn{1}{c|}{update\_target\_rate}          & 10            \\ \cline{2-9} 
\multicolumn{1}{|c|}{}                                       & \multicolumn{1}{c|}{save\_rate}          & \multicolumn{1}{c|}{20}            & \multicolumn{1}{c|}{one\_hot}             & \multicolumn{1}{c|}{true}          & \multicolumn{1}{c|}{phase}                & \multicolumn{1}{c|}{false}     & \multicolumn{1}{c|}{gamma}                         & 0.95          \\ \cline{2-9} 
\multicolumn{1}{|c|}{}                                       & \multicolumn{1}{c|}{steps}               & \multicolumn{1}{c|}{3600}          & \multicolumn{1}{c|}{test\_steps}          & \multicolumn{1}{c|}{3600}          & \multicolumn{1}{c|}{vehicle\_max}         & \multicolumn{1}{c|}{1}         & \multicolumn{1}{c|}{grad\_clip}                    & 5             \\ \cline{2-9} 
\multicolumn{1}{|c|}{}                                       & \multicolumn{1}{c|}{train\_model}        & \multicolumn{1}{c|}{false}         & \multicolumn{1}{c|}{test\_model}          & \multicolumn{1}{c|}{true}          & \multicolumn{1}{c|}{action\_interval}     & \multicolumn{1}{c|}{10}        & \multicolumn{1}{c|}{}                              &               \\ \hline
\multicolumn{9}{|c|}{\textbf{Server Parameters}}                                                                                                                                                                                                                                                                                                                                \\ \hline
\multicolumn{1}{|c|}{\textbf{pc1}}                           & \multicolumn{1}{c|}{cpu}                 & \multicolumn{3}{c|}{Intel(R) Xeon(R) Platinum 8163 CPU @ 2.50GHz}                                                   & \multicolumn{1}{c|}{cpu cores}            & \multicolumn{1}{c|}{24}        & \multicolumn{1}{c|}{mem total}                     & 251.55GB      \\ \hline
\multicolumn{1}{|c|}{\textbf{pc2}}                           & \multicolumn{1}{c|}{cpu}                 & \multicolumn{3}{c|}{Intel(R) Xeon(R) Platinum 8124M CPU @ 3.00GHz}                                                  & \multicolumn{1}{c|}{cpu cores}            & \multicolumn{1}{c|}{18}        & \multicolumn{1}{c|}{mem total}                     & 251.54GB      \\ \hline
\multicolumn{9}{|c|}{\textbf{Training Parameters}}                                                                                                                                                                                                                                                                                                                              \\ \hline
\multicolumn{1}{|c|}{thread}                                 & \multicolumn{1}{c|}{4}                   & \multicolumn{1}{c|}{ngpu}          & \multicolumn{1}{c|}{-1}                   & \multicolumn{2}{c|}{action\_pattern}                                           & \multicolumn{1}{c|}{"set"}     & \multicolumn{1}{c|}{if\_gui}                        & true          \\ \hline
\multicolumn{1}{|c|}{debug}                                  & \multicolumn{1}{c|}{false}               & \multicolumn{1}{c|}{interval}      & \multicolumn{1}{c|}{1}                    & \multicolumn{2}{c|}{savereplay}                                                & \multicolumn{1}{c|}{true}      & \multicolumn{1}{c|}{rltrafficlight}                & true          \\ \hline
\end{tabular}
}
\end{table}

\end{document}